\documentclass[10pt,journal,compsoc]{IEEEtran}
\newcommand*{\GENERALCOMPILE}{}  
\newcommand*{\HIGHRESOLUTIONFIGURE}{}  
\newcommand*{\SHOWSUPPLEMENT}{}  

\newif\ifenablearxivbuild
\enablearxivbuildtrue  

\usepackage[dvipsnames]{xcolor}
\usepackage{graphicx}
\usepackage{amsmath}
\usepackage{amsfonts}
\usepackage{pifont}
\usepackage{xspace}
\usepackage{makecell}
\usepackage{multirow}
\usepackage[export]{adjustbox}
\usepackage{threeparttable}
\usepackage{tabularx}
\usepackage{bibentry} 
\usepackage{rotating}
\usepackage[caption=false]{subfig}
\usepackage{pdflscape} 
\usepackage[citecolor=blue, colorlinks]{hyperref}  
\usepackage{tikz}
\usepackage{tikzducks}
\usetikzlibrary{spy, backgrounds}
\usetikzlibrary{positioning}

\definecolor{pending}{RGB}{51, 102, 255}
\definecolor{edited-color}{RGB}{0, 102, 255}

\newcolumntype{Y}{>{\centering\arraybackslash}X}
\newcolumntype{L}[1]{>{\raggedright\let\newline\\\arraybackslash\hspace{0pt}}m{#1}}
\newcolumntype{C}[1]{>{\centering\let\newline\\\arraybackslash\hspace{0pt}}m{#1}}
\newcolumntype{R}[1]{>{\raggedleft\let\newline\\\arraybackslash\hspace{0pt}}m{#1}}

\newcommand{\yh}{\textcolor{black}}

\newcommand{\comment}[1]{}

\newcommand*{\eg}{e.g.\@\xspace}
\newcommand*{\ie}{i.e.\@\xspace}
\newcommand*{\etal}{et~al.}

\makeatletter
\newcommand*{\etc}{%
    \@ifnextchar{.}%
        {etc}%
        {etc.\@\xspace}%
}
\makeatother

\ifCLASSOPTIONcompsoc
  \usepackage[nocompress]{cite}
\else
  \usepackage{cite}
\fi
\ifCLASSINFOpdf
\else
\fi
\begin{document}

%
\title{PU-Flow: a Point Cloud Upsampling Network\\ with Normalizing Flows}
%
%
%
%

\author{Aihua Mao,
        Zihui Du,
        Junhui Hou,~\IEEEmembership{Senior Member,~IEEE,}
        Yaqi Duan,
        Yong-jin Liu,~\IEEEmembership{Senior Member,~IEEE,}
        Ying He
\IEEEcompsocitemizethanks{
\IEEEcompsocthanksitem A. Mao, Z. Du, Y. Duan are with School of Computer Science and Engineering, South China University of Technology, Guangzhou, China. E-mails: ahmao@scut.edu.cn, \{csusami,csduanyaqi\}@mail.scut.edu.cn.
\IEEEcompsocthanksitem J. Hou is with the Department of Computer
Science, City University of Hong Kong, Hong Kong. E-mail: jh.hou@cityu.edu.hk.
\IEEEcompsocthanksitem Y.-J. Liu is with BNRist, MOE-Key Laboratory of Pervasive Computing,
Department of Computer Science and Technology, Tsinghua University, Beijing, China. E-mail: liuyongjin@tsinghua.edu.cn.
\IEEEcompsocthanksitem Y. He is with School of Computer Science and Engineering, Nanyang Technological University, Singapore. E-mail: yhe@ntu.edu.sg.
}
\thanks{Manuscript received April 19, 2005; revised August 26, 2015. \\ 
(Corresponding authors: Aihua Mao and Junhui Hou)}
}

%
%

\markboth{Journal of \LaTeX\ Class Files,~Vol.~14, No.~8, August~2015}%
{Shell \MakeLowercase{\textit{et al.}}: Bare Demo of IEEEtran.cls for Computer Society Journals}
%



\IEEEtitleabstractindextext{%
\begin{abstract}
Point cloud upsampling aims to generate dense point clouds from given sparse ones, which is a challenging task due to the irregular and unordered nature of point sets.
To address this issue, we present a novel deep learning-based model, called PU-Flow, which incorporates normalizing flows and weight prediction techniques to produce dense points uniformly distributed on the underlying surface.
Specifically, we exploit the invertible characteristics of normalizing flows to transform points between Euclidean and latent spaces and formulate the upsampling process as ensemble of neighbouring points in a latent space, where the ensemble weights are adaptively learned from local geometric context.
Extensive experiments show that our method is competitive and, in most test cases, it outperforms state-of-the-art methods in terms of reconstruction quality, proximity-to-surface accuracy, and computation efﬁciency.
The source code will be publicly available at \url{https://github.com/unknownue/pu-flow}.
\end{abstract}

}

\maketitle

\IEEEdisplaynontitleabstractindextext

%
\IEEEpeerreviewmaketitle

\IEEEraisesectionheading{\section{Introduction}\label{sec:introduction}}

\IEEEPARstart{P}{oint} clouds, as one of the most accessible 3D data formats, have been used in a wide range of scenarios, including geometric analysis, robotic object detection and autonomous driving.
With compact storage and flexible organization in representing diverse 3D objects of complex structures and geometry, point clouds have attracted increasing research interest.
However, raw points produced from LiDAR sensors or Depth cameras are often sparse, noisy and non-uniform due to hardware limitation of 3D scanning devices.
Many 3D analysis tasks, such as robotic perception, point rendering, and surface reconstruction, highly depend on the quality of input point clouds.

Therefore, point cloud upsampling, the ability to generate dense points from sparse input, is required for point cloud analysis.
Early optimization-based methods \cite{alexa2003computing}, \cite{lipman2007parameterization}, \cite{huang2013edge} use shape priors to guide point generation.
These methods work on smooth and well-distributed points but have difficulty in processing more complex geometry.

In recent years, deep neural networks (DNNs) have brought new insights into point cloud upsampling in a data-driven manner.
Yu \etal \cite{Yu2018PUNetPC} first introduced PU-Net to extract embedding features from multi-scale patches and expands upsampled points by multi-branch \emph{multi-layer perceptrons} (MLPs).
As the first end-to-end point cloud upsampling network, PU-Net \cite{Yu2018PUNetPC} demonstrates the feasibility of learning-based methods.
Thereafter, many representative approaches, including EC-Net \cite{yu2018ec}, MPU \cite{Wang2019PatchBasedP3}, PU-GAN \cite{li2019pugan}, PUGeo-Net \cite{qian2020pugeo} \yh{and MAFU \cite{QianFlexiblePU2021}}, have been proposed to further improve the quality of point generation.
However, these methods generate points simply through coordinate reconstruction.
They may overemphasize the coordinate similarity between the sparse input points and ground-truth while omitting the underlying distribution of the model surface.
Besides, as the upsampling factor is binding to the point encoding/decoding process \cite{Yu2018PUNetPC}, \cite{Wang2019PatchBasedP3}, \cite{li2019pugan}, numerous parameters are required to offer variation for duplicated features (to avoid clustering effect) and preserve uniformity.

\begin{figure}[t]
\centering
\includegraphics[width=\linewidth]{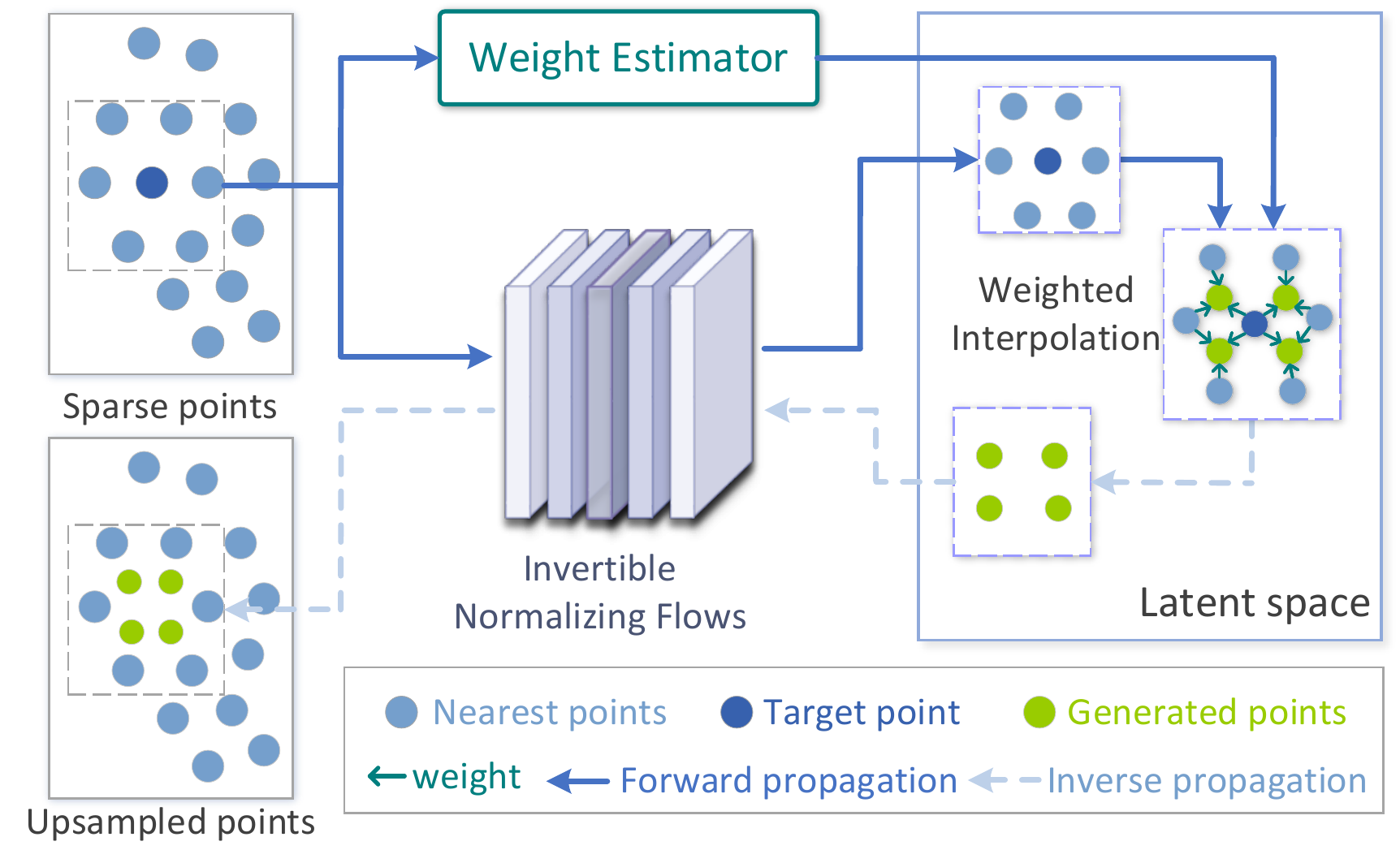}
\caption{
A schematic illustration of our method.
Input sparse patch (blue points) are first transformed into latent distribution $p_{\vartheta}(z)$ by forward propagation of normalizing flows.
Then, we produce new latent points $\hat{z}$ (green points) by weighted interpolation of existing neighbors, where weights (green arrows) are predicted by a weight estimator.
Finally, we transform $\hat{z}$ back into Euclidean space by inverse propagation and obtain upsampled patch.
For simplicity, we only show this process for a single point.
}
\label{fig:illustration-principle}
\end{figure}

\yh{
In this study, we present a generative pipeline for point cloud upsampling.
Similar to image super-resolution techniques \cite{hu2019meta}, \cite{chen2021learning}, we produce new points by weighted interpolation among local neighboring points.
Particularly, there are two modules in our pipeline, the point transformer and the weight estimator.
The point transformer formulates the transformation of point features between Euclidean space and latent space by leveraging normalizing flows (NFs).
We propose to perform weighted interpolation in latent space, with the weights adaptively predicted by the weight estimator, as illustrated in Fig. \ref{fig:illustration-principle}.
}

\yh{
NFs are known to be an invertible generative framework, which parametrizes a bijective mapping of a simple distribution into a more complex distribution.
Thus, arbitrary manipulations in latent space reflects a bijective change in Euclidean space.
By taking advantage of the invertibility of flows, we formulate the point encoding and decoding processes into a shared network.
Through this way, we do not need a specific decoder for coordinate reconstruction like previous works, which helps to avoid the reconstruction error and reduce network parameters.
}

\yh{
Previous works generally expand points by feature replication, which may lead to cluster phenomenon (\ie non-uniformity) in practise.
By contrast, our method upsamples points by adaptively interpolating local neighbors under a prior distribution, where point variations are naturally introduced during the interpolation process.
Therefore, there is no longer need to design extra modules to ensure point diversity, such as code assignment \cite{Wang2019PatchBasedP3}, \cite{li2019pugan} and multi-branch MLPs \cite{Yu2018PUNetPC}.
}

\yh{
Our upsampling pipeline is designed to formulate the point transformation and weight estimation processes into two separated branches.
On one hand, this design decouples the functionality of the point transformer and weight estimator and thus simplifies the optimization goal for each sub-network.
On the other hand, it disentangles the task of expanding points from point decoding, such that the upsampling factor is not bind to the point transformation process.
}





In summary, the contributions of this paper are as follows:

\begin{itemize}
\item We innovatively formulate the 3D point cloud upsampling problem from the perspective of learned local interpolation in a latent space.
\item \yh{We present a new upsampling pipeline, which cooperates the NFs and weight estimation techniques.
By exploiting the invertibility of NFs, this pipeline ensures bijective mapping of point sets between coordinates and their latent representation.}
\item Through qualitative and quantitative evaluations on both synthetic and real-scanned datasets, we demonstrate the advantages of our PU-Flow over state-of-the-art works.
\end{itemize}

\section{Related Works}

\subsection{Optimization-based upsampling methods}
A number of optimization-based methods for point cloud upsampling or consolidation have been proposed over the past decade.
Alexa \etal \cite{alexa2003computing} computed a Voronor diagram on underlying surface by \emph{moving least squares} methods, and generated points at vertices of the diagram.
Lipman \cite{lipman2007parameterization} introduced \emph{locally optimal projection} (LOP) operator, a parametrization-free approach, for point resampling and surface reconstruction. 
Successively, Huang \etal \cite{huang2009consolidation} designed a weighted LOP operator to further consolidate the ability to handle sharp edges, outliers, and non-uniformity.
Although the aforementioned methods can achieve good results, they are limited in processing smooth surfaces.
Later, Huang \etal \cite{huang2013edge} developed an \emph{edge-aware resampling} (EAR) method to progressively resample point set as well as approaching edge singularities, while its resampling effect is highly dependent on the accuracy of normal estimation.
To complete large missing regions, Wu \etal \cite{wu2015deep} presented a consolidation method by introducing a \emph{deep} representation for points.

In summary, optimization-based methods generally require geometric priors (\eg, normal) or assumption of smooth distribution, which restrict their application scope.
In contrast, deep learning-based methods have more powerfull generalization ability without manual parameters tuning for different point sets.

\subsection{Deep learning-based upsampling methods}
In recent years, deep learning has been widely used in many fields of point clouds learning, including classification \cite{qi2017pointnet}, \cite{qi2017pointnet++}, \cite{wang2019dynamic}, \cite{li2018so}, \cite{liu2019relation}, segmentation \cite{te2018rgcnn}, \cite{wang2019graph}, \cite{shimin2020}, registration \cite{elbaz20173d}, \cite{aoki2019pointnetlk}, \cite{CTFNet21}, denoising \cite{zhang2020pointfilter}, \cite{zhongwei2021}, generation \cite{yushi2021}, completion \cite{yuan2018pcn}, \cite{tchapmi2019topnet}, \cite{huang2020pf}, visualization \cite{chen2019b}, \cite{arnaud2020}, \etc
As the pioneer in applying neural networks to point cloud analysis, PointNet \cite{qi2017pointnet} and PointNet++ \cite{qi2017pointnet++} propose to use shared MLP and symmetric functions as feature extractor.

Based on the architecture of PointNet++, Yu \etal \cite{Yu2018PUNetPC} presented the first end-to-end deep learning framework, namely PU-Net, for point cloud upsampling.
PU-Net \cite{Yu2018PUNetPC} extracts different hierarchical features from multi-scale patches, and upsampling points are generated from multi-branch MLPs by coordinate reconstruction.
It is optimized by a joint loss function including reconstruction and repulsion loss. PU-Net outperforms previous optimization-based methods, but the upsampling results still suffer from the cluster phenomenon and lack of ﬁne-grained structure.
Subsequently, Yu \etal \cite{yu2018ec} extended PU-Net with edge-aware loss function to consolidate edge smoothness.
Yifan \etal \cite{Wang2019PatchBasedP3} proposed a progressive upsampling network with dense connection and feature interpolation operator to bridge the upsampling unit on different levels.
This network can adapt a large upsampling factor (\eg, 16$\times$) by progressively feeding points to 2x upsampling unit.
However, this mechanism requires step-by-step training for each unit, which is not ﬂexible for tuning a large upsampling factor in practice.
Subsequently, Li \etal \cite{li2019pugan} developed a generative adversarial network called PU-GAN, as well as a uniform metric to supervise upsampling quality. PU-GAN uses a self-attention unit to enhance feature integration and expand point features through the up–down–up pattern.
Although PU-GAN \cite{li2019pugan} achieves impressive results on non-uniform data, there is no guarantee that the network can converge to the best performance stably in each training.
Qian \etal \cite{qian2020pugeo} proposed PUGeo-Net to learn the intrinsic features of local geometry.
They applied normal estimation to refine coordinate correction.
Recently, Qian \etal \cite{Qian_2021_CVPR} proposed PU-GCN, a graph-based convolutional approach.
The introduced NodeShuffle module can be integrated into existing upsampling pipeline.
Li \etal \cite{li2021dispu} introduced Dis-PU, which incorporates a dense generator to generate a rough point set and a spatial refiner module to employ offset refinement.

\yh{
Our work is more related to MAFU \cite{QianFlexiblePU2021}, which extends the idea of linear interpolation to create new points and employs a flexible training strategy to enable a flexible upsampling factor. Specifically,
based on the linear approximation theory, MAFU \cite{QianFlexiblePU2021} interpolates the coordinates of neighbouring points and predicts point-wise offsets to reduce high-order approximation errors.
By contrast, our method performs interpolation in a latent space instead of the Euclidean space and is not required to refine coordinates.
}

Compared to previous works, our method unify the point encoding and decoding processes into a shared framework.
By decoupling the task of of point transformer and weight estimator, the point transformation process
is independent on the upsampling factor.

\subsection{Normalizing Flows in Point Cloud Learning}
Compared with familiar generative models, such as VAE and GAN, NFs represent another generative family that has existed for a long time but only becomes popular in recent years.
NFs have many types of implementation, such as planar flows \cite{rezende2015variational}, \cite{berg2018sylvester}, autoregressive models \cite{kingma2016improving}, \cite{oliva2018transformation}, coupling-based flows \cite{dinh2014nice}, \cite{dinh2016density}, \cite{kingma2018glow}, and continuous flows \cite{grathwohl2018ffjord}, \cite{yang2019pointflow}.
Dinh \etal \cite{dinh2014nice} introduced a coupling method to enable highly expressive transformations for flows, and this idea is further improved in \cite{dinh2016density}, \cite{kingma2018glow}, \cite{durkan2019neural}.
With coupling architecture, one can apply arbitrary complex convolutions in inference, and the calculation of the Jacobian is extensively simplified.
In recent years, continuous normalizing flows (CNF) has become a promising innovation on flow models.
Based on the Neural ODE solver \cite{chen2018neuralode}, CNF can achieve competitive performance to discrete flows \cite{kingma2018glow} but with fewer parameters.

The flow-based methods have been successfully adapted to a wide range of generative scenarios, such as image generation \cite{ardizzone2019guided}, \cite{lugmayr2020srflow}, cross-domain learning \cite{pumarola2020c}, \cite{grover2019alignflow}, video prediction \cite{kumar2019videoflow}, graph generation \cite{liu2019graph}, and audio synthesis \cite{prenger2019waveglow}, \cite{kim2018flowavenet}.
It is natural to generalize this idea to point clouds generation tasks.
Yang \etal \cite{yang2019pointflow} presented PointFlow, a CNF-based framework, to learn a two-level hierarchy of distributions of given shapes.
As the first flow-based network in point cloud learning, PointFlow can be further extends to a variety of applications \cite{rempe2020caspr}, \cite{abdal2020styleflow}.
Then, Pumarola \etal \cite{pumarola2020c} proposed C-Flow to explore the potential of bridging different domains (\eg, image and point clouds) with NFs.
Klokov \etal \cite{klokov20eccv} proposed a discrete PointFlow network to alleviate the slow convergence and difficult training issue of PointFlow \cite{yang2019pointflow}.
Postels \etal \cite{postels2021go} introduced the mixture model of NFs and showed the improved generative ability to single NF model.
These works mainly concentrate on improving the flow-based generative capability.
However, there are few works paying attention to real-world applications in point cloud analysis.

In this study, we take advantage of the invertible capacity of NFs to transform point clouds between Euclidean and latent spaces.
To the best of our knowledge, no prior work has applied NFs to point cloud upsampling tasks.

\section{Method}

\subsection{Overview}

Given a sparse point set $\mathcal{P}=\left\{p_{i} \in \mathbb{R}^{D}\right\}_{i=1}^{N}$, our goal is to predict a dense point set $\hat{\mathcal{X}}=\left\{\hat x_{i} \in \mathbb{R}^D\right\}_{i=1}^{R\times N}$, where $N$ is the number of points and $R$ is upsampling factor.
In this study, we only consider the coordinate of point attributes with $D=3$.
The generated point set $\hat{\mathcal{X}}$ is expected to meet the following requirements:

\begin{itemize}
    \item $\hat{\mathcal{X}}$ should retain the geometric details represented by $\mathcal P$, while $\mathcal P$ is not necessary to be a subset of $\hat{\mathcal{X}}$.
    \item $\hat{\mathcal{X}}$ should be complete and uniformly distributed in both local and global areas.
\end{itemize}

In this study, we propose to utilize NFs to model the mapping of the point distribution between Euclidean space and latent space, which enables us to formulate the point cloud upsampling as the problem of learning point interpolation in latent space, as illustrated in Fig. \ref{fig:illustration-principle}.
Specifically, given an input sparse point set $\mathcal P$, we first convert it to latent variable $z=f(\mathcal{P})$ with an invertible transformation defined by NFs.
Then, we interpolate points in $z$ and obtain dense latent variable $\hat{z}^{R}$, where the interpolation weights are learned from the point-wise local neighbours.
Finally we transform $\hat{z}^R$ to a dense point cloud $\hat{\mathcal{X}}$ by the inverse mapping $\hat{\mathcal {X}}=f^{-1}(\hat{z}^{R})$.


\begin{figure*}[t]
\centering
\includegraphics[width=\linewidth]{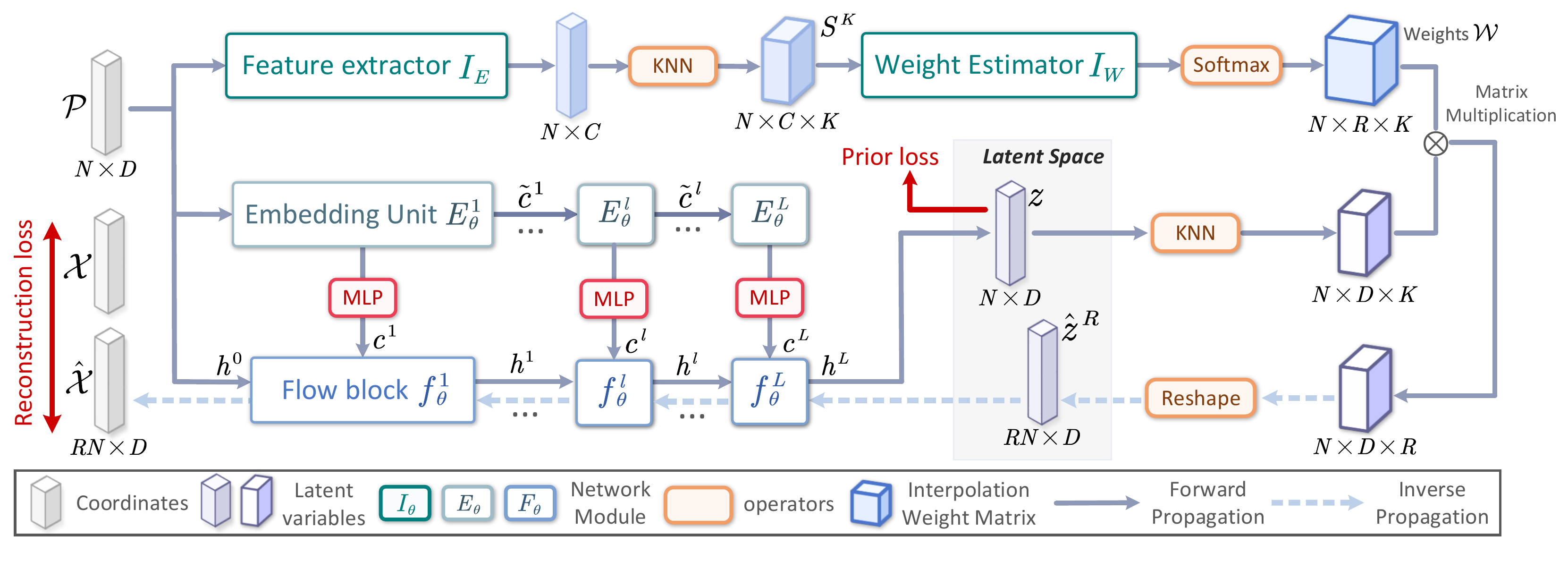}
\vspace{-0.30in}
\caption{
Network architecture of the proposed method.
Given a sparse patch $\mathcal P$ with $N$ points, our method transforms $\mathcal P$ into latent distribution $z$ by a sequence of flow blocks $f_{\theta}^l$ (forward propagation).
Flow block $f_{\theta}^l$ is conditioning on features $c^l$, which is generated by point embedding unit $E_{\theta}^l$.
Then interpolation module $I_{\theta}$ predicts neighbour weights $\mathcal{W}$ by analyzing local context of $k$-nearest graph for each point.
By interpolating latent variable $z$, we get dense latent distribution $\hat{z}^R$, with upsampling factor $R$.
Upsampled patch is generated by converting $\hat{z}^R$ to Euclidean space by inverse flow $F_{\theta}^{-1}$ (inverse propagation).
The network is trained end-to-end by minimizing prior loss $\mathcal {L}_{\text{prior}}$ of $\mathcal{P}$ and reconstruction loss $\mathcal{L}_{\text{rec}}$.
}
\label{fig:illustration-network-arch}
\end{figure*}

\subsection{Flow-based Upsampling Method}
\label{section:method_normalizing_flows}
A normalizing flow is a series of invertible transformations of distribution.
It is generally used to model an intractable, complex distribution by a simple prior distribution.
Formally, let $z \in \mathbb{R}^{N\times D}$ be a latent variable of base distribution $p_{\vartheta}(z)$ with the known density, \ie, $z \sim p_{\vartheta}(z)$.
Given a dataset of observations $\mathcal P$, we aim to learn an invertible transformation $f_{\theta}(\cdot)$ to parameterize mapping from $\mathcal P$ to tractable density $p_\vartheta(z)$:
\begin{equation}\label{formula_nf_forward}
    z = f_{\theta}(\mathcal P;\mathcal C),
\end{equation}
where $\mathcal C=\psi\left(\mathcal P\right)$, and $\psi(\cdot)$ is an arbitrary function that extracts conditional features from $\mathcal P$.
Here, we refer to $f_{\theta}$ as conditional normalizing flows, which is generally parameterized by a neural network with parameters $\theta$.
Note that, $f_{\theta}$ is required to be a bijective transformation, which indicates that the dimension of points remains unchanged during distribution transforms.

By exploring the geometric structure in local context of each point, we apply weighted interpolation over the $k$-nearest neighbors, producing the upsampled latent variables $\hat{z}^{R}\in\mathbb{R}^{RN\times D}$
\begin{equation}
    \hat{z}^{R}_i=I_{\theta}(z_i, \mathcal{N}(p_i)),
\end{equation}
where $\mathcal{N}(p_i)$ denotes the $k$-nearest neighbors of latent point $p_i$ and $I_{\theta}$ represents the interpolation function.
Given conditional features $\mathcal C$ and latent points $\hat{z}^R$, the inverse mapping $g_{\theta}(\cdot) = f_{\theta}^{-1}(\cdot)$ implicitly defines the point decoding process
\begin{equation}\label{formula_nf_inverse}
    \hat{\mathcal{X}} = g_{\theta}(\hat{z}^R;\mathcal C),
\end{equation}
where $\hat{\mathcal{X}}$ is an upsampled estimation of $\mathcal P$.
In contrast to decoding point by MLPs, utilizing the inverse mapping $f_{\theta}^{-1}$ can take advantage of the invertibility of NFs and help to reduce the reconstruction error and the number of network parameters.

As the single-layer flow model has limited non-linear capabilities, in practice, the flow network $f_{\theta}$ is composed of a sequence of $L$ invertible layers.
Let $h^l$ be the output of the $l$-th flow layers, then $h^{l+1}$ is defined as
\begin{equation}\label{formula_nf_layer_forward}
    h^{l+1}=f^{l+1}_{\theta}(h^l;\mathcal{C}^l),
\end{equation}
where $f^{l}_{\theta}$ is the $l$-th flow layer, $h^0=\mathcal P$, $h^L=z$, and $\mathcal C^l$ is the corresponding conditional features at the $l$-th layer.
With the change of variable formula \cite{dinh2016density} and the chain rule, the probability density of the given input $\mathcal P$ can be computed as
\begin{equation}
\label{formula_log_prob}
\begin{aligned}
\log p\left(\mathcal P\mid\mathcal C,\theta\right)=\log p_{\vartheta}\left(f_{\theta}\left(\mathcal{P} ; \mathcal{C}\right)\right)+\log\left|\operatorname{det}\frac{\partial f_{\theta}}{\partial \mathcal{P}}(\mathcal{P} ; \mathcal{C})\right|\\
=\log p_{\vartheta}\left(f_{\theta}\left(\mathcal{P} ; \mathcal{C}\right)\right)+\sum_{l=1}^L \log \left|\operatorname{det} \frac{\partial f_{\theta}^{l}}{\partial h^l}(h^{l} ; \mathcal C^l)\right|,
\end{aligned}
\end{equation}
where the term $\left|\operatorname{det} \frac{\partial f_{\theta}}{\partial \mathcal{P}}(\mathcal{P};
\mathcal{C})\right|$ is the Jacobian determinant of transformation $f_\theta$, measuring the volume changing \cite{dinh2014nice} caused by $f_\theta$.
Generally, $f_{\theta}$ is trained by maximum likelihood principle with gradient descent techniques.

\section{Network Architecture}
\label{section:network_arch}

According to the idea in Section \ref{section:method_normalizing_flows}, the overall architecture of PU-Flow depicted in Fig. \ref{fig:illustration-network-arch} includes an invertible flow module $F_{\theta}$ (Section \ref{section:network-flow-module}) and a point interpolation module $I_{\theta}$ (Section \ref{section:network_interpolation}).
Furthermore, to enhance the transformation capability of the flow layer, we introduce a hierarchical point embedding module $E_{\theta}$ (Section \ref{section:network_point_embedding}) to augment conditional features (\ie $\mathcal C$ in Section \ref{section:method_normalizing_flows}) for $F_{\theta}$ in both forward and inverse propagation.

\begin{figure*}[t]
\centering
\includegraphics[width=0.95\linewidth]{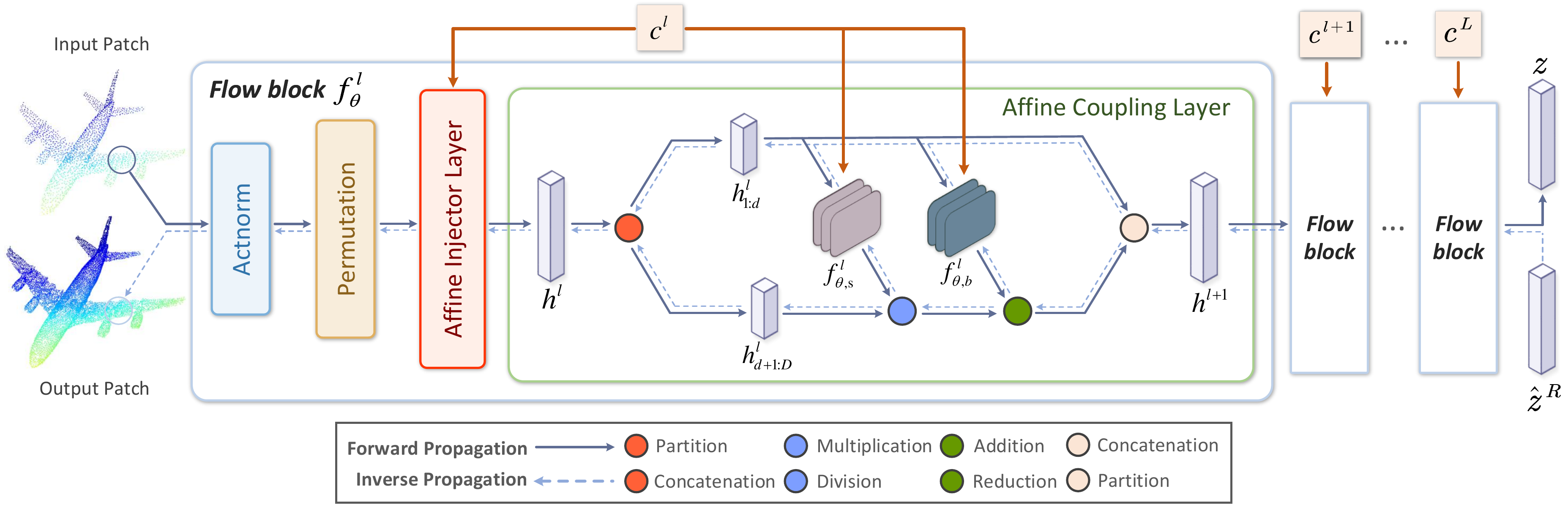}
\vspace{-0.05in}
\caption{
Illustration of discrete flow block $f_{\theta}^l$ and conditional affine coupling layer.
Each flow block consists of 4 flow components.
The affine coupling/injector layer are conditioning on embedding features $c^l$ from $E_{\theta}^l$ (Section \ref{section:network_point_embedding}) in both forward and inverse propagation.
}
\label{fig:illustration-flow-block}
\end{figure*}

\subsection{Hierarchical Point Embedding}
\label{section:network_point_embedding}

\textbf{Dimensional bottleneck.}
NFs are designed to ensure analytical invertibility.
This fact poses a challenge that each flow component must output the same dimensionality as the input data (the dimension of raw point clouds is only $D=3$).
This constraint conflicts with the widely adopted intelligence of deep learning that learns features with a higher dimension than that of input data, resulting in limited transform capability of each flow block.
This issue can be referred to as the \emph{dimensional bottleneck} problem.

To alleviate this limitation, previous works, such as RealNVP \cite{dinh2016density} and Glow \cite{kingma2018glow}, propose to increase flows depth or use a multi-scale architecture coupled with squeezing operation.
Simply increasing the depth of flows requires a large amount of parameters, leading to slow training speed and decreased training stability.
Meanwhile, squeezing operator exchanges feature channel with spatial dimension, which is mainly designed for image manipulation.
However, it is non-trivial to adopt squeezing to point cloud processing due to the unorder nature of a point set.

\noindent
\textbf{Hierarchical Embedding.}
Based on the above analysis, we propose a parallel sequence of embedding units to augment additional point-wise features for flow block $f_{\theta}^l$ as follows:
\begin{equation}\label{formula:point-embedding-unit}
    \tilde{c}^l=E_{\theta}^l(\tilde{c}^{l-1}),
\end{equation}
where $\tilde{c}^l\in \mathbb{R}^{N\times C}$ denotes the high-level features outputted by $l$-th unit $E_{\theta}^l$, and $\tilde{c}^0=\mathcal P$.
As $E_{\theta}$ is not a component of the NFs, it does not need to be invertible and can be arbitrary flexible architectures.
One can consider the hierarchical embedding as a pattern of feature fusion.

In this study, the point embedding unit $E_{\theta}^l$ is constructed by a stack of densely connected graph convolutional layers \cite{wang2019dynamic}, where the neighbour size of the graph is fixed to 16.
It utilizes dense connections to enable richer contextual vision of multi-scale features.
A more detailed description of unit $E_{\theta}^l$ can be found in supplementary material.

We further employ a simple MLP layer to obtain a specialized point-wise conditional features $\tilde{c}^l$ for each individual flow block (\ie $c^l=\text{MLP}(\tilde{c}^l)$), as shown in Figure \ref{fig:illustration-network-arch}.
Note that the embedding features $\mathcal C=\{c^l\}_{l=1}^L$ are shared in both forward and inverse propagation and only need to be computed once.
We investigate the impact of embedding unit $E_{\theta}^l$ in Section \ref{sec:ablation-study}.

\subsection{Flow Module}
\label{section:network-flow-module}

During the forward propagation, the flow module $F_{\theta}$ accepts input patch $\mathcal P=h^0$, perform transformation through a sequence of flow blocks, and output latent variables $z=h^L$ from final block $f^L_{\theta}$.
Through exact log-likelihood training, $F_{\theta}$ learns a conditional mapping from $\mathcal P$ to latent distribution $p_{\vartheta}(z)$.


To be specific, $F_{\theta}$ is composed of $L$ blocks.
Each flow block consists of four flow layers, including actnorm \cite{kingma2018glow}, permutation layer \cite{kingma2018glow}, affine injector layer \cite{lugmayr2020srflow} and affine coupling layer \cite{dinh2016density}.
We carefully design these layers to satisfy the invertible requirement (see supplementary material for detailed formulation).
Among these layers, the affine coupling/injector layers merge features $c^l$ from point embedding unit $E_{\theta}^l$ and employ distribution transformation to intermediate point representation $h^l$, as shown in Fig. \ref{fig:illustration-flow-block}.


During the inverse propagation, interpolated latent variable $\hat{z}^{R}=h^L$ is fed into last flow block $f^L_{\theta}$ as input, and upsampled estimation $\hat{\mathcal{X}}$ is generated through inverse flow pass $F_{\theta}^{-1}$.
We duplicate features in $c^l$ to match the number of points between $c^l$ and $h^l$ during conditioning.

It is worth pointing out that the flow block $f_{\theta}^l$ can also be implemented using the continuous flow block (\ie ODE-Net \cite{chen2018neuralode}).

\subsection{Interpolation Module}
\label{section:network_interpolation}


Given a sparse point set $\mathcal{P}$ as input, we first enrich point-wise features $s_i$ by a feature extractor $I_{E}$:
\begin{equation}\label{formula-interp-feat-extractor}
    \mathcal S=I_{E}(\mathcal{P}).
\end{equation}
where $\mathcal{S}=\left\{s_{i} \in \mathbb{R}^{C}\right\}_{i=1}^{N}$.
See supplementary material for detailed implementation of $I_{E}$.

To expand (upsample) latent points, we need to interpolate $R$ points for each latent point $z_i$.
Therefore, we gather a set of neighbors as point-wise local context $\mathcal{S}_{i}^{K}=\left\{s_{i}^{k}\right\}_{k=1}^{K}$ by $k$NN algorithm, where $s_i^k$ is the $k$-th nearest neighbor of $s_i$ in $\mathcal{S}$.
Then, we predict $R$ groups of weights for each latent point $z_i$ by a weight estimator $I_W$.
This process can take the form:
\begin{equation}\label{formula_interp_weightmatrix}
    \mathcal W_i=I_{W}(\mathcal{S}_{i}^{K}),
\end{equation}
where matrix $\mathcal W_i=\left[w_i^{1,1},w_i^{1,2},\ldots, w_i^{r,k},\ldots,w_i^{R,K}\right]\in \mathbb{R}^{R\times K}$ denotes the predicted weights for $K$ nearest neighbors of the $i$-th point.
The element $w_i^{r,k}$ indicates the weight of the $k$-th latent point $z_i^k$ among $k$NN in the $r$-th interpolation result.
$I_W$ is simply parameterized as MLPs.

Before interpolation, we apply a softmax function to the generated weight
\begin{equation}\label{formula_interp_softmax}
    \tilde{w}_{i}^{r,k}=\frac{e^{w_{i}^{r,k}}}{\sum_{k=1}^{K} e^{w_{i}^{r,k}}},
\end{equation}
such that we obtain normalized weights that satisfy $\tilde{w}_i^{r,k}\geq 0$ and $\sum_{k=1}^{K} \tilde{w}_{i}^{r,k}=1$.

Finally, interpolation can be formulated as matrix multiplication performed on latent variable $z_i$
\begin{equation}\label{formula_interp_interp}
\hat{z}_{i}^{r}=\sum_{z^k\in \mathcal N\left(p_i\right)} \tilde{w}_{i}^{r, k} z^{k},
\end{equation}
where $\hat{z}_{i}^{r} \in \mathbb{R}^{D}$ is the interpolated point of the $r$-th result, and $\mathcal N\left(p_i\right)$ is the set of latent variables in $k$NN field of point $p_i$.
Note that the neighbor relationship is constructed in Euclidean space to keep consistent neighborship with conditional features $c$ in $E_{\theta}$.

After interpolation, we flatten interpolated dense points $\hat{z}^R$ and feed them into the inverse propagation pass of flow module $F_{\theta}$, as shown in the dotted path of Fig. \ref{fig:illustration-network-arch}.

\begin{table*}[t]
\caption{Quantitative comparisons of state-of-the-art methods on PU1K ($N=2048$). We highlight the best and second best results  in bold and underline, respectively.}
\centering
\begin{tabular}{c||c|ccccc||c|ccccc}
\hline
Ratio & \multicolumn{6}{c||}{$R=4$} & \multicolumn{6}{c}{$R=16$} \\ \hline
\thead{Methods} & \thead{Network \\ Size} & \thead{CD \\ \scalebox{0.88}[0.88]{($10^{-4}$)}} & \thead{EMD \\ \scalebox{0.88}[0.88]{($10^{-2}$)}} & \thead{HD \\ \scalebox{0.88}[0.88]{($10^{-2}$)}} & \thead{P2F \\ \scalebox{0.88}[0.88]{($10^{-3}$)}} & \thead{JSD \\ \scalebox{0.88}[0.88]{($10^{-2}$)}} & \thead{Network \\ Size} & \thead{CD \\ \scalebox{0.88}[0.88]{($10^{-4}$)}} & \thead{EMD \\ \scalebox{0.88}[0.88]{($10^{-2}$)}} & \thead{HD \\ \scalebox{0.88}[0.88]{($10^{-2}$)}} & \thead{P2F \\ \scalebox{0.88}[0.88]{($10^{-3}$)}} & \thead{JSD \\ \scalebox{0.88}[0.88]{($10^{-2}$)}} \\
\hline
EAR \cite{huang2013edge} & - & 9.198 & 4.572 & 0.770 & 2.029 & 11.31 & - & 9.021 & 5.134 & 0.896 & 2.712 & 12.16 \\
PU-Net \cite{Yu2018PUNetPC} & 10.1 MB & 6.654 & 3.872 & 6.927 & 5.747 & 8.847 & 24.5 MB & 6.767 & 4.470 & 1.156 & 4.625 & 11.66 \\
MPU \cite{Wang2019PatchBasedP3} & 23.1 MB & 4.401 & 3.100 & 0.340 & 1.237 & 5.204 & 92.5 MB & 4.203 & 3.421 & 0.454 & 1.456 & 6.105 \\
PU-GAN \cite{li2019pugan} & 9.5 MB & 4.239 & 3.036 & 0.540 & 1.580 & 5.382 & 9.5 MB & 4.365 & 3.826 & 0.724 & 2.164 & 7.827 \\
PUGeo-Net \cite{qian2020pugeo} & 26.6 MB & 3.695 & 2.809 & 0.325 & \underline{1.189} & \underline{4.267} & 26.7 MB & 2.791 & 3.208 & 0.386 & 1.359 & 4.896 \\
PU-GCN \cite{Qian_2021_CVPR} & 1.8 MB & 3.847 & 2.862 & 0.374 & 1.385 & 4.509 & 1.8 MB & 2.875 & 3.243 & 0.471 & 1.507 & 5.244 \\
Dis-PU \cite{li2021dispu} & 13.2 MB & 3.941 & 2.897 & 0.362 & 1.336 & 4.568 & 13.2 MB & 2.987 & \underline{3.178} & 0.430 & 1.469 & 5.157 \\
MAFU \cite{QianFlexiblePU2021} & 4.7 MB & 3.624 & \underline{2.776} & 0.450 & {\bf 1.139} & 4.365 & 4.7 MB & 2.745 & 3.219 & 0.487 & {\bf 1.303} & 5.087 \\
\hline
Ours (discrete) & 3.4 MB & \underline{3.613} & 2.861 & {\bf 0.310} & 1.324 & 4.311 & 3.4 MB & \underline{2.675} & 3.217 & {\bf 0.367} & 1.416 & \underline{4.864} \\
Ours (continuous) & 3.3 MB & {\bf 3.563} & {\bf 2.741} & \underline{0.321} & 1.315 & {\bf 4.021} & 3.3 MB & {\bf 2.548} & {\bf 3.114} & \underline{0.373} & \underline{1.350} & {\bf 4.728} \\
\hline
\end{tabular}
\label{tab:quantitative_pu1k}
\end{table*}%

\subsection{Training Objects}

Let $\mathcal{P}\in \mathbb{R}^{N\times D}$ and $\mathcal{X}\in \mathbb{R}^{RN\times D}$ be the sets of sparse input and ground-truth dense points, respectively, with upsampling factor $R$.
We design a joint loss function to train PU-Flow in an end-to-end manner.
This objective function consists of two components: the reconstruction loss to encourage the generated points $\mathcal {\hat X}$ and reference $\mathcal{X}$ to share the same distribution, and the prior loss to optimize the transformation capability of flow module $F_{\theta}$ by maximizing the likelihood of observation $\mathcal P$.

\textbf{Reconstruction loss.}
We employ Earth Mover’s distance (EMD) loss to measure the similarity between $\mathcal{X}$ and $\mathcal {\hat X}$:
\begin{equation}\label{formula_loss_EMD}
    \mathcal {L}_{\text{rec}}=\mathcal{L}_\text{EMD}(\mathcal {\hat X},\mathcal{X})=\min _{\phi: \mathcal {\hat X} \rightarrow \mathcal X} \sum_{x_{i} \in \mathcal {\hat X}}\left\|x_{i}-\phi\left(x_{i}\right)\right\|_{2},
\end{equation}
where $\phi: \mathcal {\hat X} \rightarrow \mathcal X$ is a bijective mapping.

\textbf{Prior likelihood.}
Eq. (\ref{formula_log_prob}) allows us to train the flow layers by minimizing the negative log-likelihood (NLL) with input patch $\mathcal P$:
\begin{equation}\label{formula_loss_prior}
    \mathcal {L}_{\text{prior}}\left(\mathcal P\right)= \mathcal{L}(\mathcal{P}, \mathcal{C}; \theta)=-\log p(\mathcal{P} \mid \mathcal{C}, \theta),
\end{equation}
where $\mathcal C=E_{\theta}\left(\mathcal P\right)$.
Optimizing prior likelihood of $\mathcal P$ encourages the encoded shape representation to gain high probability under the predefined prior $p_{\vartheta}(z)$, which is modeled by the flow module $F_{\theta}$.

In our experiment, the prior $p_{\vartheta}(z)$ is simply set as standard Gaussian distribution $\mathcal N\left(0, \mathbf{I}\right)$.
In addition, $p_{\vartheta}(z)$ can also be set to Gaussian with learnable mean and variance, but we do not observe an obvious influence to model performance.

\textbf{Total loss.} 
Combining the preceding formulas, we train PU-Flow with respect to parameters $\theta$ by minimizing
\begin{equation}\label{formula_loss_joint}
    \mathcal L\left(\theta\right)=\alpha\mathcal{L}_{\text{rec}}+\beta\mathcal {L}_{\text{prior}},
\end{equation}
where $\alpha$ and $\beta$ are hyper-parameters that balance the terms.

\section{Experiments}

\subsection{Experimental setup}
\label{sec:experimental-setup}

\textbf{Datasets.}
For quantitative comparison, we train and evaluate our method on following datasets:

\begin{itemize}
\item \textbf{PU1K.}
This dataset consists of models from PU-GAN \cite{li2019pugan} and ShapeNetCore \cite{shapenet2015} of various categories, which are used in PU-GCN \cite{Qian_2021_CVPR}.
PU1K contains 1020 meshes for training and 127 meshes for evaluation.
\item \textbf{PUGeo-Net dataset.}
This dataset includes elaborate statues from Sketchfab \cite{sketchfab_website}, provided by PUGeo-Net \cite{qian2020pugeo}.
It contains 90 high-resolution meshes for training and 13 for testing, with complex geometry and high-frequency details.
\item \textbf{PU36.}
To achieve more generalized evaluation results of more categories, we constructed a new dataset for evaluation, containing 36 models collected from Sketchfab \cite{sketchfab_website}.
Please refer to supplementary material for gallery of all shapes.
\item \textbf{PU-GAN dataset.}
\yh{
This dataset \cite{li2019pugan} includes 120 models for training and 27 for testing.
The testing set contains a variety of basic shapes.
}
\item \textbf{FAMOUSTHINGI.}
This dataset includes models chosen from Thingi10k \cite{zhou2016thingi10k} and PCPNet \cite{GuerreroEtAl_PCPNet} datasets, with a total of 37 shapes for evaluation.
We use FAMOUSTHINGI \cite{Rakotosaona_2021_CVPR} dataset to evaluate the quality of surface reconstruction results.
\end{itemize}

\yh{
In the experiments, the results on PU1K, PUGeo-Net, PU36 and FAMOUSTHINGI datasets are trained and evaluated on uniform data, while the results on PU-GAN dataset are trained and evaluated on non-uniform data.
The results on the PU1K, PU36 and PU-GAN datasets are evaluated by the same evaluation script as PU-GCN \cite{Qian_2021_CVPR}.
The results on the PUGeo-Net dataset is evaluated by the same evaluation script as PUGeo-Net \cite{qian2020pugeo}.
}

\begin{figure*}[t]
\hspace*{\fill}  
\begin{tikzpicture}
\node (fig) at (current page.east) {\includegraphics[width=0.20\textwidth]{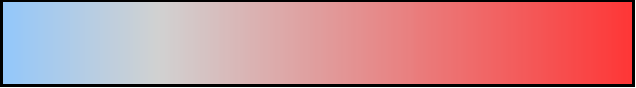}};
\node[left=0cm of fig] {Low};
\node[right=0cm of fig] {High};
\end{tikzpicture}

\hspace*{0.01\linewidth}
\ifdefined\GENERALCOMPILE
    \ifdefined\HIGHRESOLUTIONFIGURE
        \includegraphics[width=0.985\linewidth]{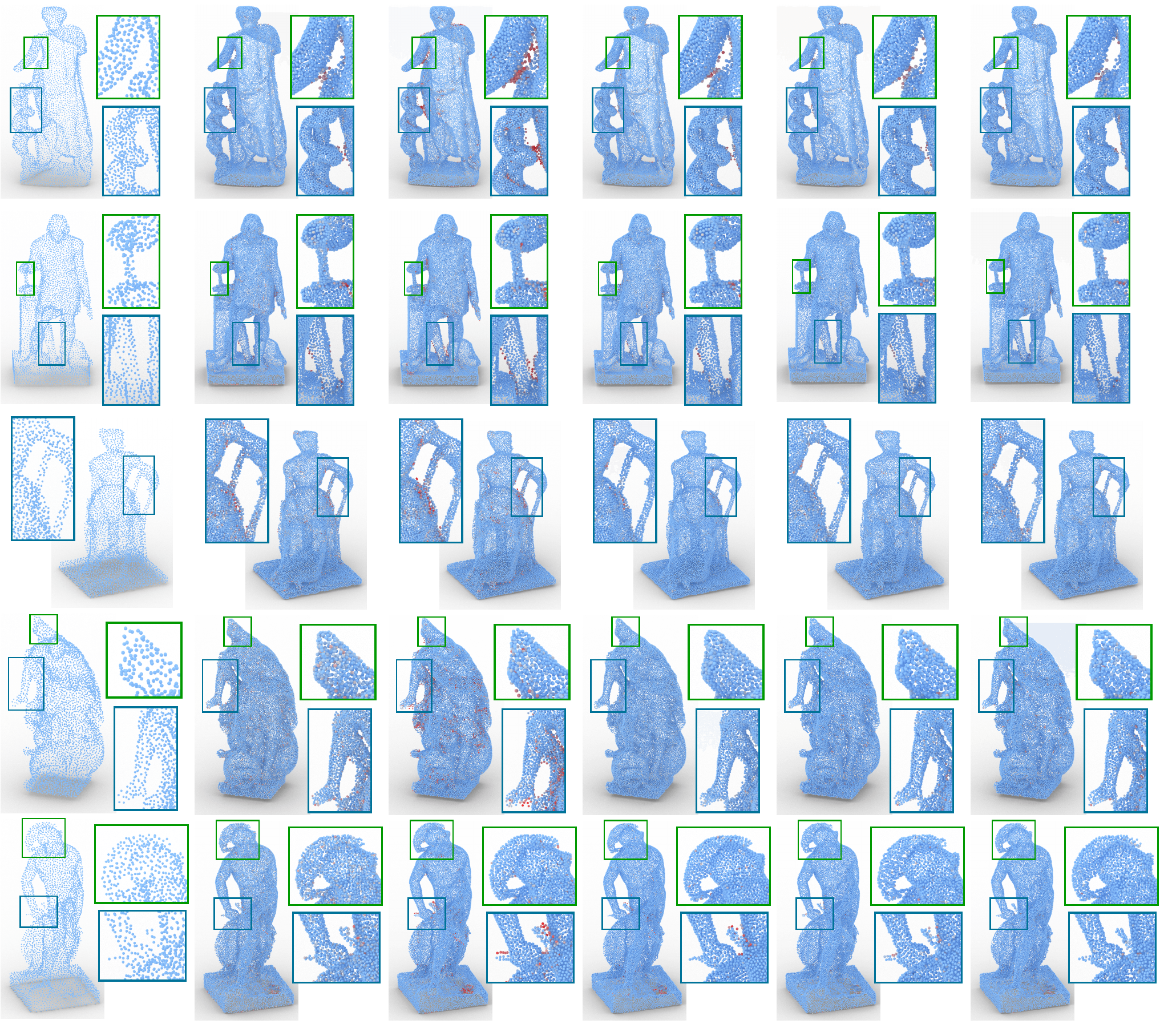}
    \else
        \includegraphics[width=0.985\linewidth]{images/plotting/reduce/sota-points-visual-comparison-reduce.pdf}
    \fi
\fi
\begin{tabularx}{\linewidth}{@{}Y@{}Y@{}Y@{}Y@{}Y@{}Y@{}}
(a) Input &
(b) PU-GCN \cite{Qian_2021_CVPR} &
(c) Dis-PU \cite{li2021dispu} &
(d) MAFU \cite{QianFlexiblePU2021} &
(e) PUGeo-Net \cite{qian2020pugeo} &
(f) Ours (discrete) \\
\end{tabularx}
\caption{
Visual Comparisons of various methods (b-f).
We visualized the P2F errors by colors for each point. The images are best viewed on screen when zoomed in.
See also the supplementary material for more comparisons and visual results.
}
\label{fig:visual-sota-points}
\end{figure*}

\noindent
\textbf{Methods under comparison.}
We compare our model with a representative optimization-based method and six state-of-the-art deep learning-based methods, including EAR \cite{huang2013edge}, PU-Net \cite{Yu2018PUNetPC}, MPU \cite{Wang2019PatchBasedP3}, PU-GAN \cite{li2019pugan}, PUGeo-Net \cite{qian2020pugeo}, PU-GCN \cite{Qian_2021_CVPR}, Dis-PU \cite{li2021dispu} \yh{and MAFU \cite{QianFlexiblePU2021}}.
For fair comparison, we use the public released code and retrain the models for all deep learning-based methods on each datasets for evaluation. \yh{
Note that as PU1K and PU-GAN datasets do not contain the normal information of points, we retrained PUGeo-Net and MAFU without using their normal generation modules.
}

\noindent
\textbf{Implementation details.}
We implemented PU-Flow with PyTorch framework.
The corresponding source code will be published later, including both discrete and continuous implementations.
The training settings, detailed network architectures, hyper-parameters and evaluation practices are provided in supplementary material.


\begin{table*}[t]
\caption{Quantitative comparisons of state-of-the-art methods on PUGeo-Net dataset and PU36 ($N=5000$,$R=4$). We highlight the best and second best results  in bold and underline, respectively.}
\centering
\begin{tabular}{c||ccccc||ccccc}
\hline
Dataset & \multicolumn{5}{c||}{PUGeo-Net dataset} & \multicolumn{5}{c}{PU36} \\
\hline
\thead{Methods} & \thead{CD \\ \scalebox{0.88}[0.88]{($10^{-2}$)}} & \thead{EMD \\ \scalebox{0.88}[0.88]{($10^{-2}$)}} & \thead{HD \\ \scalebox{0.88}[0.88]{($10^{-2}$)}} & \thead{P2F \\ \scalebox{0.88}[0.88]{($10^{-3}$)}} & \thead{JSD \\ \scalebox{0.88}[0.88]{($10^{-2}$)}} & \thead{CD \\ \scalebox{0.88}[0.88]{($10^{-4}$)}} & \thead{EMD \\ \scalebox{0.88}[0.88]{($10^{-2}$)}} & \thead{HD \\ \scalebox{0.88}[0.88]{($10^{-3}$)}} & \thead{P2F \\ \scalebox{0.88}[0.88]{($10^{-3}$)}} & \thead{JSD \\ \scalebox{0.88}[0.88]{($10^{-2}$)}} \\
\hline
EAR \cite{huang2013edge} & 0.660 & 2.745 & 1.990 & 1.489 & 1.192 & 1.695 & 2.323 & 1.571 & 1.843 & 2.319 \\
PU-Net \cite{Yu2018PUNetPC} & 0.658 & 2.419 & 1.003 & 1.532 & 0.950 & 3.245 & 2.706 & 4.505 & 9.704 & 3.960 \\
MPU \cite{Wang2019PatchBasedP3} & 0.573 & 1.491 & 1.073 & 0.808 & 0.614 & 1.049 & 1.632 & 1.040 & 1.507 & 0.895 \\
PU-GAN \cite{li2019pugan} & 0.586 & 1.957 & 1.422 & 0.889 & 0.579 & 1.516 & 2.146 & 3.624 & 2.292 & 1.926 \\
PUGeo-Net \cite{qian2020pugeo} & 0.558 & 1.479 & {\bf 0.934} & 0.617 & 0.444 & 1.016 & 1.557 & {\bf 0.909} & 1.514 & 0.832 \\
PU-GCN \cite{Qian_2021_CVPR} & 0.568 & 1.538 & 1.093 & 0.754 & 0.542 & 1.082 & 1.641 & 1.729 & 1.682 & 1.014 \\
Dis-PU \cite{li2021dispu} & 0.585 & 1.719 & 1.318 & 0.819 & 0.615 & 1.235 & 1.823 & 2.568 & 2.083 & 1.265 \\
MAFU \cite{QianFlexiblePU2021} & 0.556 & 1.480 & 1.077 & 0.653 & 0.453 & 1.060 & 1.558 & 1.024 & 1.576 & 0.974 \\
\hline
Ours (discrete) & {\bf 0.551} & {\bf 1.441} & \underline{0.951} & {\bf 0.582} & {\bf 0.422} & \underline{0.979} & {\bf 1.520} & 1.011 & {\bf 1.434} & {\bf 0.749} \\
Ours (continuous) & \underline{0.553} & \underline{1.464} & 0.963 & \underline{0.590} & \underline{0.433} & {\bf 0.973} & \underline{1.532} & \underline{0.991} & \underline{1.463} & \underline{0.781} \\
\hline
\end{tabular}
\label{tab:quantitative_pugeo}
\end{table*}%

\begin{table}[t]
\caption{Quantitative comparisons of state-of-the-art methods with non-uniform inputs on the PU-GAN dataset ($N=2048$, $R=4$). We highlight the best and second best results in bold and underline, respectively.}
\centering
\begin{tabular}{c|ccccc}
\hline
Methods & \thead{CD \\ \scalebox{0.88}[0.88]{($10^{-4}$)}} & \thead{EMD \\ \scalebox{0.88}[0.88]{($10^{-2}$)}} & \thead{HD \\ \scalebox{0.88}[0.88]{($10^{-3}$)}} & \thead{P2F \\ \scalebox{0.88}[0.88]{($10^{-3}$)}} & \thead{JSD \\ \scalebox{0.88}[0.88]{($10^{-2}$)}} \\
\hline
EAR \cite{huang2013edge} & 5.543 & 4.562 & 9.042 & 7.038 & 6.872 \\
PU-Net \cite{Yu2018PUNetPC} & 4.642 & 4.035 & 8.249 & 8.348 & 9.267 \\
MPU \cite{Wang2019PatchBasedP3} & 3.410 & 3.439 & 4.758 & 2.935 & 4.898 \\
PU-GAN \cite{li2019pugan} & {\bf 2.985} & {\bf 2.787} & 5.284 & 2.818 & {\bf 3.811} \\
PUGeo-Net \cite{qian2020pugeo} & 3.286 & 3.298 & 5.693 & \underline{2.589} & 4.500 \\
PU-GCN \cite{Qian_2021_CVPR} & 3.476 & 3.433 & 4.821 & 2.809 & 5.459 \\
Dis-PU \cite{li2021dispu} & 3.288 & \underline{3.084} & 5.425 & {\bf 2.460} & 4.479 \\
MAFU \cite{QianFlexiblePU2021} & 3.192 & 3.247 & {\bf 4.472} & 2.671 & 4.268 \\
\hline
Ours (discrete) & \underline{3.103} & 3.143 & \underline{4.631} & 2.724 & \underline{4.149} \\
\hline
\end{tabular}
\label{tab:quantitative_pugan}
\end{table}%

\begin{figure*}[t]
\centering
\ifdefined\GENERALCOMPILE
    \ifdefined\HIGHRESOLUTIONFIGURE
        \includegraphics[width=\linewidth]{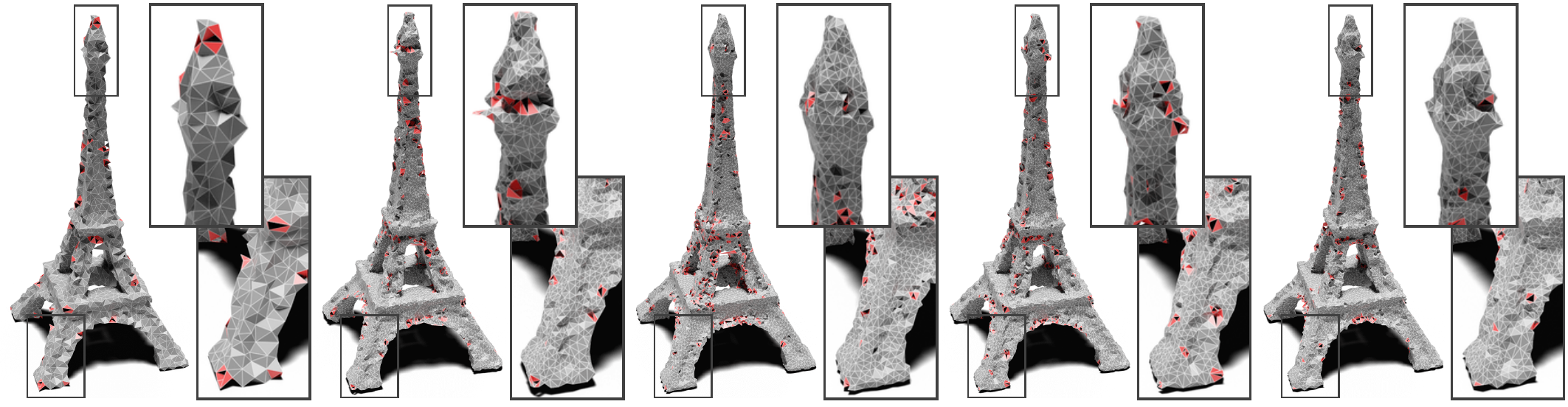}
    \else
        \includegraphics[width=\linewidth]{images/plotting/reduce/Eiffel_Tower_mini.pdf}
    \fi
\fi
\begin{tabularx}{\linewidth}{@{}Y@{}Y@{}Y@{}Y@{}Y@{}}
0.376/3.07\%/20.4 &
0.377/6.42\%/24.6 &
0.384/7.85\%/25.0 &
0.358/3.61\%/21.2 &
0.366/2.92\%/22.2 \\
\end{tabularx}
\ifdefined\GENERALCOMPILE
    \ifdefined\HIGHRESOLUTIONFIGURE
        \includegraphics[width=\linewidth]{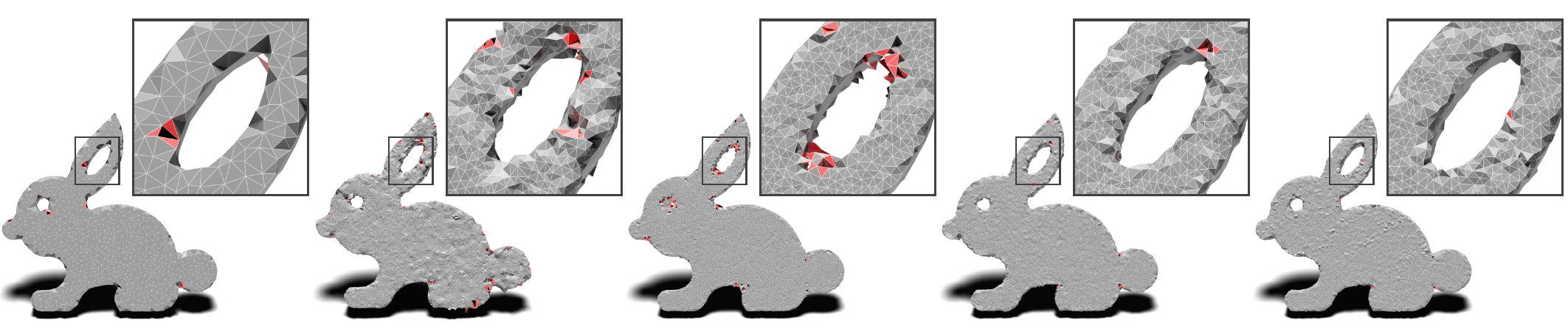}
    \else
        \includegraphics[width=\linewidth]{images/plotting/reduce/Easter_Bunny_001_Ready_to_print.pdf}
    \fi
\fi
\begin{tabularx}{\linewidth}{@{}Y@{}Y@{}Y@{}Y@{}Y@{}}
0.256/0.65\%/5.96 &
0.278/1.42\%/8.45 &
0.259/1.31\%/6.42 &
0.255/0.52\%/6.92 &
0.256/0.34\%/6.88 \\
\end{tabularx}
\ifdefined\GENERALCOMPILE
    \ifdefined\HIGHRESOLUTIONFIGURE
        \includegraphics[width=\linewidth]{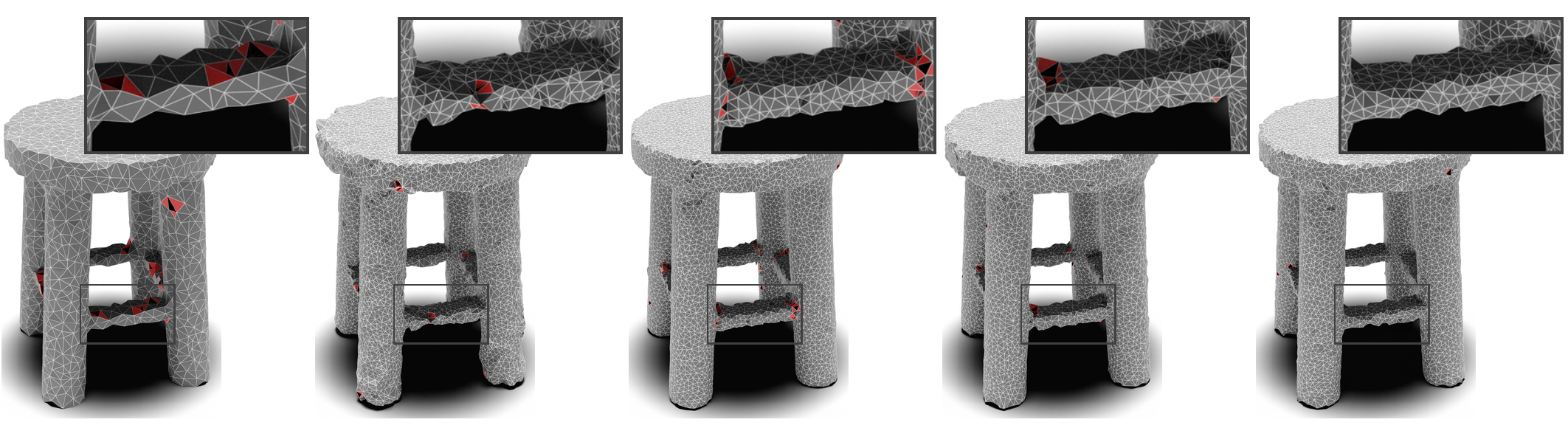}
    \else
        \includegraphics[width=\linewidth]{images/plotting/reduce/stool.pdf}
    \fi
\fi
\begin{tabularx}{\linewidth}{@{}Y@{}Y@{}Y@{}Y@{}Y@{}}
0.446/1.46\%/7.32 &
0.453/0.88\%/7.85 &
0.405/0.85\%/7.33 &
0.411/0.25\%/7.64 &
0.408/0.12\%/7.08 \\
\end{tabularx}
\vskip 0.04in

\begin{tabularx}{\linewidth}{@{}Y@{}Y@{}Y@{}Y@{}Y@{}}
(a) Input &
(b) PU-GCN \cite{Qian_2021_CVPR} &
(c) Dis-PU \cite{li2021dispu} &
(d) PUGeo-Net \cite{qian2020pugeo} &
(e) Ours (discrete) \\
\end{tabularx}

\vspace*{-0.08in}
\caption{
Visual Comparisons of reconstructed surfaces from upsampled points of various methods (b-e).
The first column shows the mesh reconstructed from sparse input points.
To visualize the artifacts of mesh surface, we mark the non-manifold triangles in red.
We display the Chamfer distance (multiplied by 100), the percentage of non-watertight edges (NW) and normal reconstruction error (NR) below each object.
See supplementary material for more visual results.
}
\label{fig:visual-sota-mesh-1}
\end{figure*}

\subsection{Comparisons on upsampled points}
\label{rec:comparisons-on-upsampled-points}

\noindent
\textbf{Evaluation metrics.}
We employ five evaluation metrics, including (\romannumeral1) Chamfer Distance (CD), (\romannumeral2) Earth Mover’s distance (EMD), (\romannumeral3) Point-to-surface distance (P2F), (\romannumeral4) Hausdorff distance (HD) and (\romannumeral5) Jensen-Shannon divergence (JSD).
All metrics are estimated on the whole point set after merging from upsampled patches.
The lower the values are, the better the upsampling quality is.

\noindent
\textbf{Quantitative comparison on uniform inputs.}
Table \ref{tab:quantitative_pu1k} shows the comparison results evaluated on PU1K dataset.
We also compare the network size as an intuitional metric to evaluate the memory efficiency of networks.
We can see that both our discrete and continuous models can achieve the lowest values on most evaluation metrics.

To be specific, deep learning-based methods significantly outperform EAR on all metrics.
This reveals the superiority of the deep learning technique compared with the optimization-based methods.
For PU-GAN, it fails to maintain a good estimation of the HD and P2F metrics.
For PU-GCN, we use the pretrained model provided by \cite{Qian_2021_CVPR} and achieve consistent results in \cite{Qian_2021_CVPR}.
The results in Table \ref{tab:quantitative_pu1k} show the performance of PU-GCN is inferior to other methods.
We also observe that PUGeo-Net and Dis-PU have competitive performance on all metrics, but both of them require relatively more network parameters.
Since PUGeo-Net can not be trained without normal supervision on PU1K training set, we use the pretrained model of PUGeo-Net instead.

In contrast to the aforementioned works, our method disentangles the point expanding and decoding processes, which extensively reduces the number of parameters but still preserves reasonable results.
Our method achieves the best balance between network size and generation quality.
Table \ref{tab:quantitative_pugeo} shows the results of different methods evaluated on PUGeo-Net dataset and PU36 dataset.
Still, our method maintains \comment{superior}advantages on most metrics.

\noindent
\textbf{Quantitative comparison on non-uniform inputs.}
\yh{
Table \ref{tab:quantitative_pugan} shows the results evaluated on the PU-GAN dataset.
From Table \ref{tab:quantitative_pugan}, we observe that PU-GAN \cite{li2019pugan} achieves the best results on CD, EMD and JSD metrics, but fails to preserve the superiority in terms of HD and P2F metrics.
Among other non-GAN-based methods, our method achieves competitive performance to PUGeo-Net \cite{qian2020pugeo} and MAFU\cite{QianFlexiblePU2021} and outperforms PU-GCN \cite{Qian_2021_CVPR} and Dis-PU \cite{li2021dispu}.
}

\noindent
\textbf{Qualitative comparison.}
We visualize the 4x upsampling results of different methods with 5K inputs points in Fig. \ref{fig:visual-sota-points}.
We use color to reveal the P2F error for each point.

Compared with other methods, our method achieves minimum average errors.
It can better preserve the smoothness of local regions and produce a reliable shape, while other methods tend to produce more noisy points between some complex adjacent regions, as shown in Fig. \ref{fig:visual-sota-points} (b), (c), and (d).

\begin{figure*}[t]
\centering
\includegraphics[width=\textwidth]{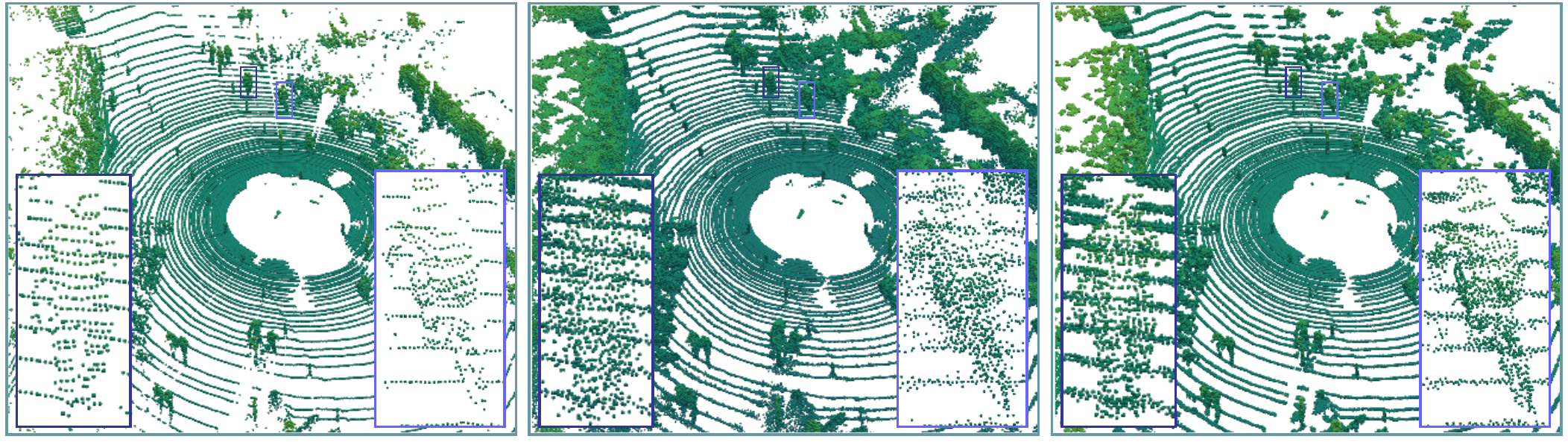}
\begin{tabularx}{\textwidth}{@{}Y@{}Y@{}Y@{}}
(a) Input & (b) PU-GCN \cite{Qian_2021_CVPR} & (c) PUGeo-Net \cite{qian2020pugeo}
\end{tabularx}
\includegraphics[width=\textwidth]{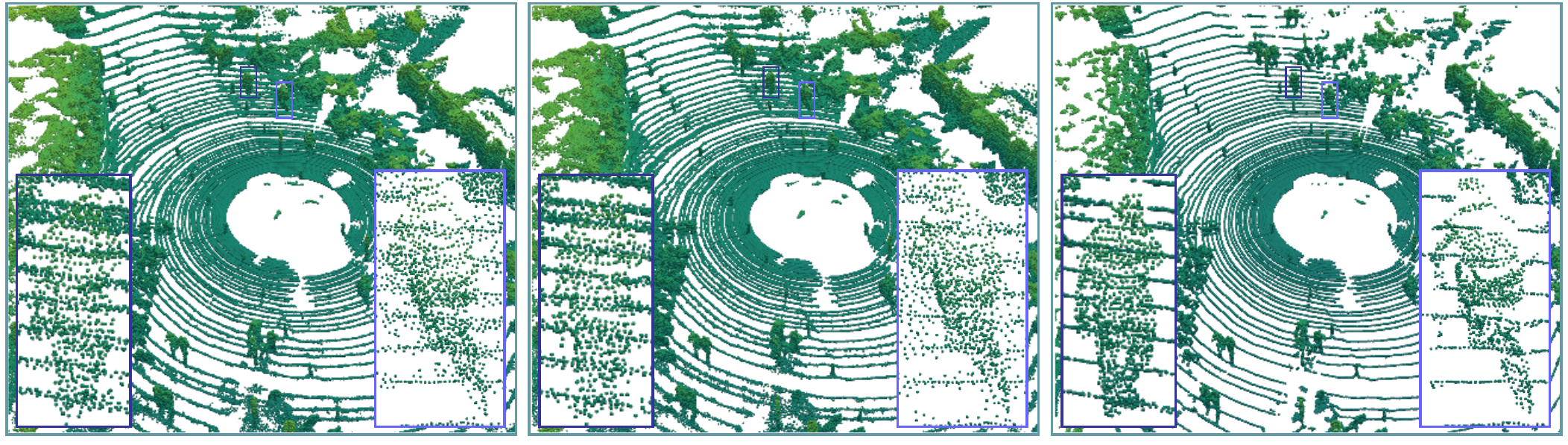}
\begin{tabularx}{\textwidth}{@{}Y@{}Y@{}Y@{}}
(d) PU-GAN \cite{li2019pugan} & (e) Dis-PU \cite{li2021dispu} & (f) Ours (discrete)
\end{tabularx}
\vspace{-0.25in}
\caption{
Visual comparisons of upsampling results ($R=4$) on the KITTI \cite{Geiger2013IJRR} dataset.
See supplementary material for more visual results.
}
\label{fig:visual-comparison-KITTI}
\end{figure*}

\subsection{Comparisons on reconstructed surface}

To further demonstrate the generation quality of our method, we compare the surface reconstruction results from upsampled points ($N=2500$, $R=4$) with state-of-the-art methods.

Specifically, we employ DSE-meshing \cite{Rakotosaona_2021_CVPR}, a cutting edge method for mesh reconstruction from point clouds, as mesh generator.
We use DSE-meshing instead of traditional methods, such as screened Poisson Sampling Reconstruction \cite{kazhdan2013screened} and ball-pivoting surface reconstruction \cite{bernardini1999ball}, for the following reasons:
(\romannumeral1) Traditional methods generally require additional information (\eg normal data) and careful parameter selection to obtain satisfactory results.
Since DSE-meshing is an end-to-end method, we can achieve more fair comparisons by eliminating the influence of normal accuracy and manual parameters tuning.
(\romannumeral2) The quality of upsampled points have significant impact to DSE-meshing.
Thus, we can employ the quantitative comparison between reconstructed mesh to evaluate the upsampling quality of various methods.


\noindent
\textbf{Evaluation metrics.}
We consider three metrics to evaluate mesh quality:
(\romannumeral1) Chamfer Distance (CD), (\romannumeral2) the percentage of non-watertight edges (NW), (\romannumeral3) normal reconstruction error in degrees (NR).
CD measures the distance between point sets sampled on reconstructed and ground truth surface.
NW counts the number of triangle edges that are only shared by one triangle.
NR measures the angle difference of normals between reconstructed and ground truth surface.
For these metrics, the lower the values are, the better the mesh quality is.


\begin{table}[t]
\caption{Quantitative comparisons of mesh reconstruction results on the FAMOUSTHINGI dataset  \cite{Rakotosaona_2021_CVPR}. We highlight the best and second best results  in bold and underline, respectively.}
\centering
\begin{tabular}{c|ccc}
\hline
Methods & CD ($10^{-2}$) & NW (\%) & NR (degree) \\
\hline
EAR \cite{huang2013edge} & 0.684 & 4.203 & 16.67 \\
PU-Net \cite{Yu2018PUNetPC} & 1.071 & 11.285 & 29.06 \\
MPU \cite{Wang2019PatchBasedP3} & 0.409 & 0.886 & 10.29 \\
PU-GAN \cite{li2019pugan} & 0.453 & 3.533 & 12.96 \\
PUGeo-Net \cite{qian2020pugeo} & {\bf 0.393} & 0.849 & \underline{9.75} \\
PU-GCN \cite{Qian_2021_CVPR} & 0.421 & 2.464 & 12.15 \\
Dis-PU \cite{li2021dispu} & 0.453 & 2.796 & 11.98 \\
MAFU \cite{QianFlexiblePU2021} & 0.407 & 0.854 & 9.87 \\
\hline
Ours (discrete) & \underline{0.394} & {\bf 0.744} & {\bf 9.69} \\
Ours (continuous) & 0.398 & \underline{0.806} & 10.01 \\
\hline
Reference & 0.326 & 0.397 & 5.22 \\
\hline
\end{tabular}
\label{tab:quantitative_famousthingi}
\end{table}%

\noindent
\textbf{Quantitative comparison.}
Table \ref{tab:quantitative_famousthingi} summarizes the comparison results evaluated on FAMOUSTHINGI dataset.
We also include the mesh reconstructed from ground truth points (denoted as Reference).
From Table \ref{tab:quantitative_famousthingi}, we can observe the similar trend with Table \ref{tab:quantitative_pugeo}.
Obvious performance gap still exists between the best one and reference.
In particular, both our discrete and continuous model yield lowest reconstruction error (CD and NR) and produce least non-manifold edges (NW).

\noindent
\textbf{Qualitative comparison.}
Fig. \ref{fig:visual-sota-mesh-1} visualizes the reconstructed mesh between representative works.
We can observe notable artifacts in the region with large curvature, especially for geometrically complex surface (\eg tower in Fig. \ref{fig:visual-sota-mesh-1}).
The uniformity and outliers of upsampled points affect the quality of mesh significantly.
PUGeo-Net and our method both achieve the most promising reconstruction quality.
However, our method produce less non-manifold (bad) triangles across most objects, demonstrating its better point distribution.


\subsection{Upsampling Real-world Data}
To confirm the robustness on more complicated and unseen data, we evaluate our method on two real-world point clouds datasets:
KITTI \cite{Geiger2013IJRR} and ScanObjectNN \cite{uy-scanobjectnn-iccv19}.

\noindent
\textbf{KITTI.}
This dataset captures the point clouds of driving scenes with its data produced by LiDAR sensors.
The raw input data severely suffer from sparsity, occlusion, and non-uniformity.
For example, some vehicles, people, and plants are only described with few points and the density of points vary across distance to center.
As shown in Fig. \ref{fig:visual-comparison-KITTI}, we observe other methods suffer from sparsity and non-uniformity issues and thus produce more outliers.
Our method can generate dense points with more fine-grained details, resulting in better object visibility when compared with other methods.

\begin{figure}[t]
\centering
\includegraphics[width=\linewidth]{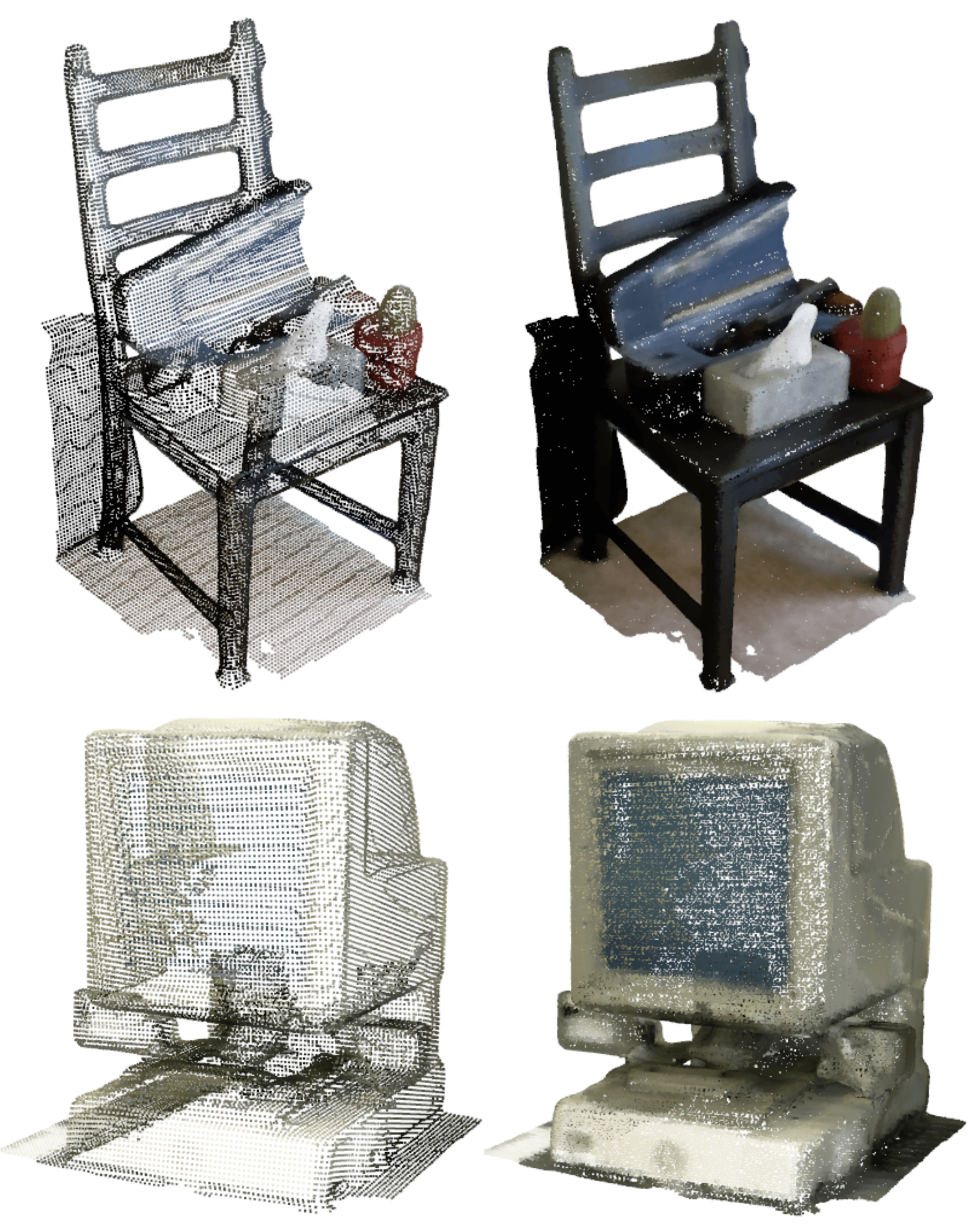}
\vspace{-0.3in}
\caption{
Upsampling results by PU-Flow in different categories (chair, computer monitor).
For better visualization, the color of upsampled points (2nd column) are assigned by the color of the closest input point (1st column).
See the supplementary material for more visual results.
}
\label{fig:scanobjectnn-part1-v1}
\end{figure}

\noindent
\textbf{ScanObjectNN.}
This dataset comprises point clouds of scanned indoor scenes, where objects are divided into 15 categories.
The raw input objects are cluttered with background and suffer from the partial occlusion and the scan line distribution pattern.
As shown in Fig. \ref{fig:scanobjectnn-part1-v1}, our method improves the visibility quality and makes objects more distinguishable from the background.

\begin{table}[t]
\caption{Ablation study of flow architectures.}
\centering
\begin{tabular}{c|l|cc}
\hline
\multicolumn{2}{c|}{Ablation} & \thead{CD \\ \scalebox{0.88}[0.88]{($10^{-4}$)}} & \thead{P2F \\ \scalebox{0.88}[0.88]{($10^{-3}$)}} \\
\hline
\multicolumn{2}{c|}{Vanilla pipeline} & 1.17 & 1.89 \\ \hline
\multirow{4}{*}{\thead{Embedding \\ Unit $E_{\theta}$}} & None & 1.31 & 2.36 \\
& PointNet \cite{qi2017pointnet} & 1.14 & 1.84 \\
& EdgeConv \cite{wang2019dynamic} & 1.03 & 1.57 \\
& PAConv \cite{xu2021paconv} & 1.09 & 1.64 \\
\hline
\multirow{3}{*}{\thead{Point \\ Decoder}} & MLP (w/ $\mathcal{L}_{\text{prior}}$) & 1.35 & 2.16 \\
& MLP (w/o $\mathcal{L}_{\text{prior}}$) & 1.12 & 1.91 \\
& xyz-interpolation & 1.38 & 1.84 \\
\hline
\multirow{3}{*}{\thead{Graph used in \\ $E_{\theta} + I_{\theta}$}} & static + dynamic & 2.05 & 2.86 \\
& dynamic + static & 1.02 & 1.45 \\
& dynamic + dynamic & 1.95 & 2.52 \\
\hline
\multicolumn{2}{c|}{\thead{Full pipeline (Point decoder: $F^{-1}_{\theta}$, \\ Graph: static + static)}} & {\bf 0.98} & {\bf 1.43} \\ \hline
\end{tabular}
\label{tab:ablation-flow-architectures}
\end{table}

\begin{figure}[t]
\centering
\includegraphics[width=\linewidth]{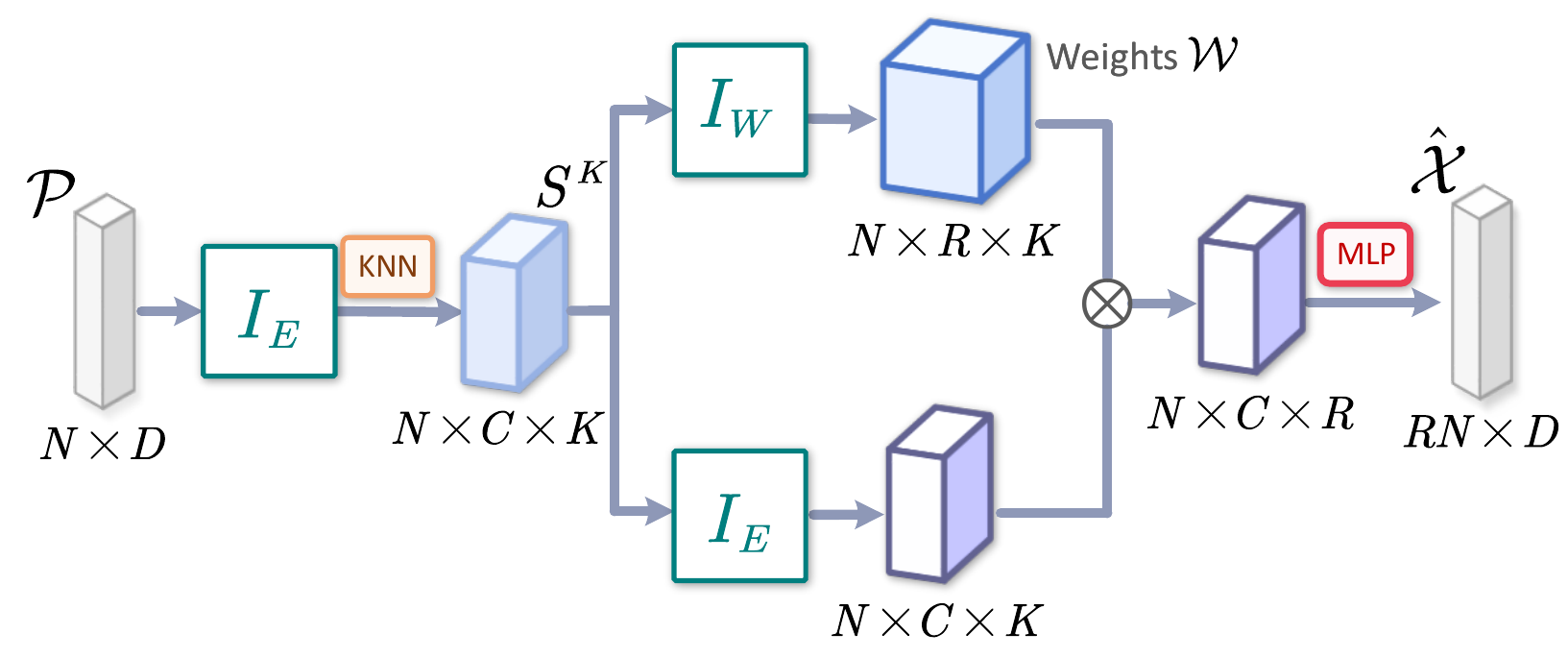}
\caption{Network architecture of the vanilla model.}
\label{fig:illustration-vanilla-arch}
\end{figure}

\subsection{Ablation Study}
\label{sec:ablation-study}

In this section, we quantitatively evaluate the contribution of network design of PU-Flow.
We use the discrete model instead of continuous model for evaluation, because they have very close performance.
The benchmarks are evaluated on PU36 dataset with input points $N=5000$ and upsampling factor $R=4$.

\begin{figure}[t]
\centering
\includegraphics[width=\linewidth]{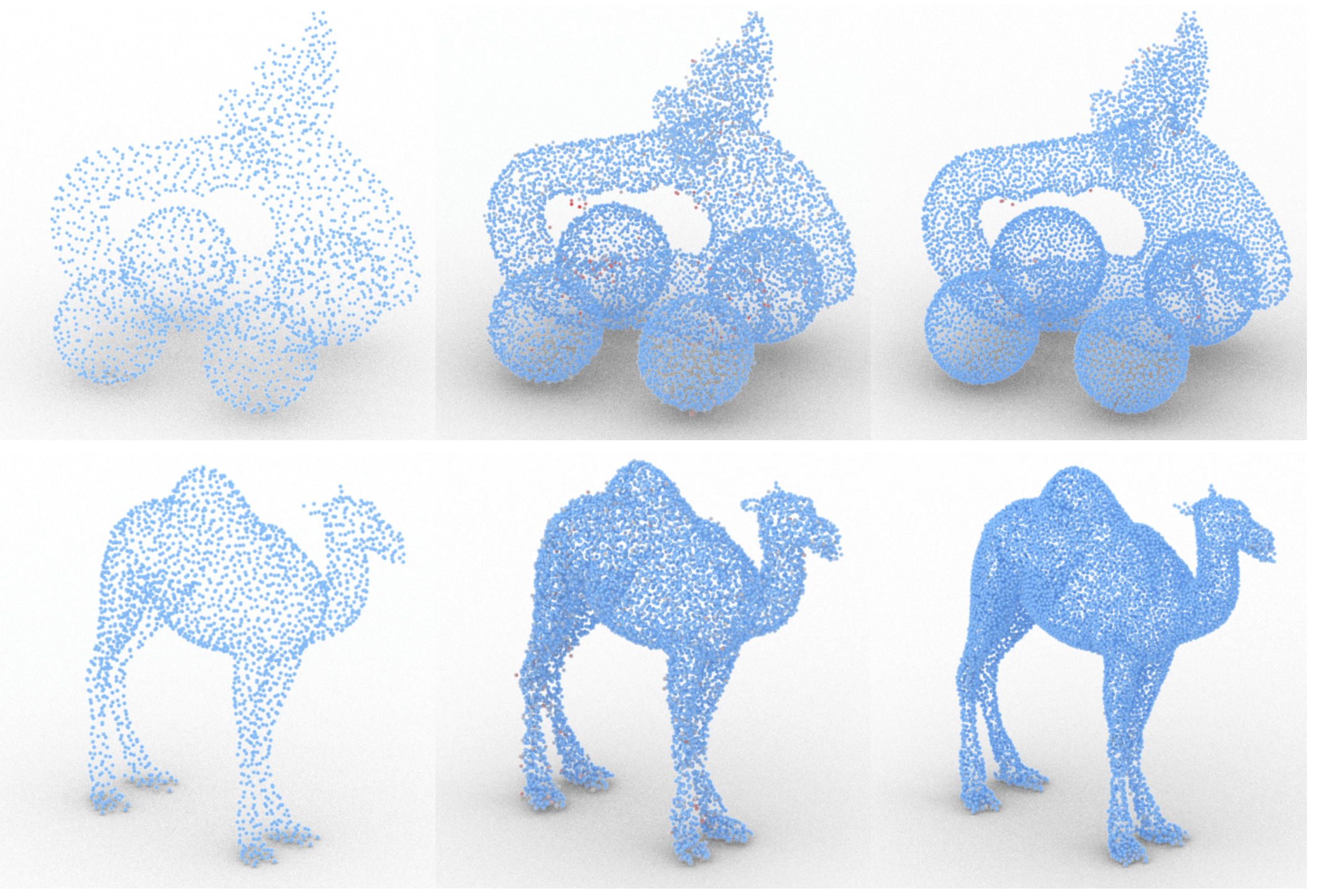}
\begin{tabularx}{\linewidth}{@{}Y@{}Y@{}Y@{}Y@{}}
Input & Vanilla & Ours \\
\end{tabularx}
\caption{Visual comparisons of generated points between the vanilla model and our method ($N=2048$, $R=4$).}
\label{fig:visual-vanilla-comp}
\end{figure}

\noindent
\textbf{Flow architecture.}
\yh{
We first construct a vanilla model (denoted as vanilla pipeline in Table \ref{tab:ablation-flow-architectures}), which implements the basic idea of weighted interpolation for upsampling.
As shown in Figure \ref{fig:illustration-vanilla-arch}, this model generates weights and high-level point abstraction from shared point-wise semantic features.
}
Compared to our full pipeline, the vanilla model has relatively low performance, demonstrating that it has difficulty to generate appropriate weights for latent features.
\yh{
Figure \ref{fig:visual-vanilla-comp} shows the upsampled results by the vanilla model and our method, where the vanilla model fails to preserve a smooth distribution on surface and produces more jitters.
}

We investigate the impact of point embedding unit $E_{\theta}$, as shown in Table \ref{tab:ablation-flow-architectures}.
The vanilla model does not use features from $E_{\theta}$ (the None row).
In this way, the flow module $F_{\theta}$ uses independent transformation for each point and thus our method suffers from underfitting.
By integrating features from modern feature extractor, $F_{\theta}$ reveals promising transform capability, thus leading to substantial performance boost.

To validate the effectiveness of generating points by inverse propagation $F_{\theta}^{-1}$, we replace it with MLPs used in previous works \cite{Yu2018PUNetPC}, \cite{Wang2019PatchBasedP3}, \cite{li2019pugan}, in which $\mathcal {L}_{\text{prior}}$ is not needed.
The results in Table \ref{tab:ablation-flow-architectures} show that using $F_{\theta}^{-1}$ for coordinate reconstruction can better preserve intricate structures than MLPs, which demonstrates the feasibility of $F_{\theta}^{-1}$.


\begin{figure}[t]
\centering
\includegraphics[width=0.9\linewidth]{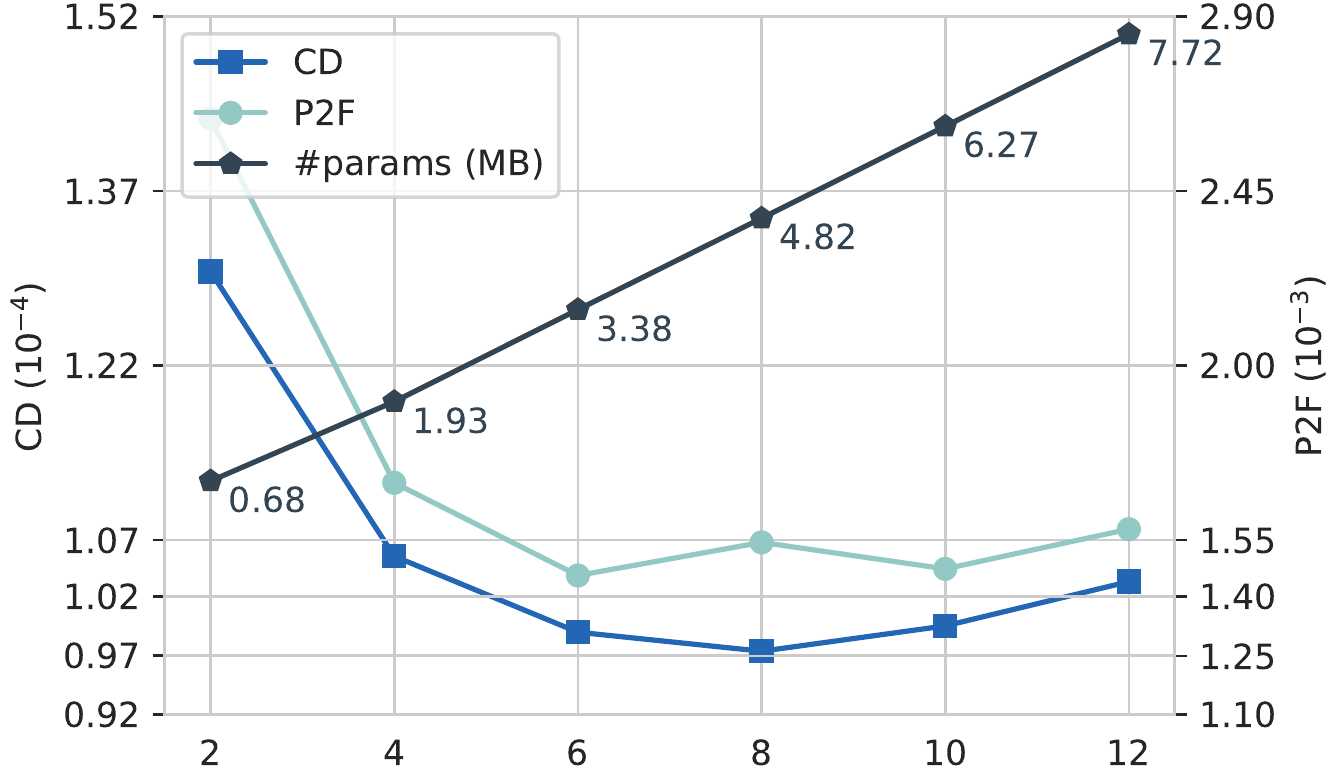}
\vspace{-0.10in}
\caption{Ablation study of number of flow block from $L=2$ to $L=12$.}
\label{fig:ablation-num-flow-blocks}
\end{figure}

Fig. \ref{fig:ablation-num-flow-blocks} shows the effect of the number of flow blocks $L$ used in PU-Flow.
When $L$ is relatively low, our method achieves better performance as $L$ increase.
When $L\geq8$, the performance gain becomes unobvious, with the cost of computational overhead and increasing network parameters.
The best performance of the full pipeline is achieved when setting $L=6$ to $L=8$.

\begin{figure}[t]
\centering
\includegraphics[width=0.9\linewidth]{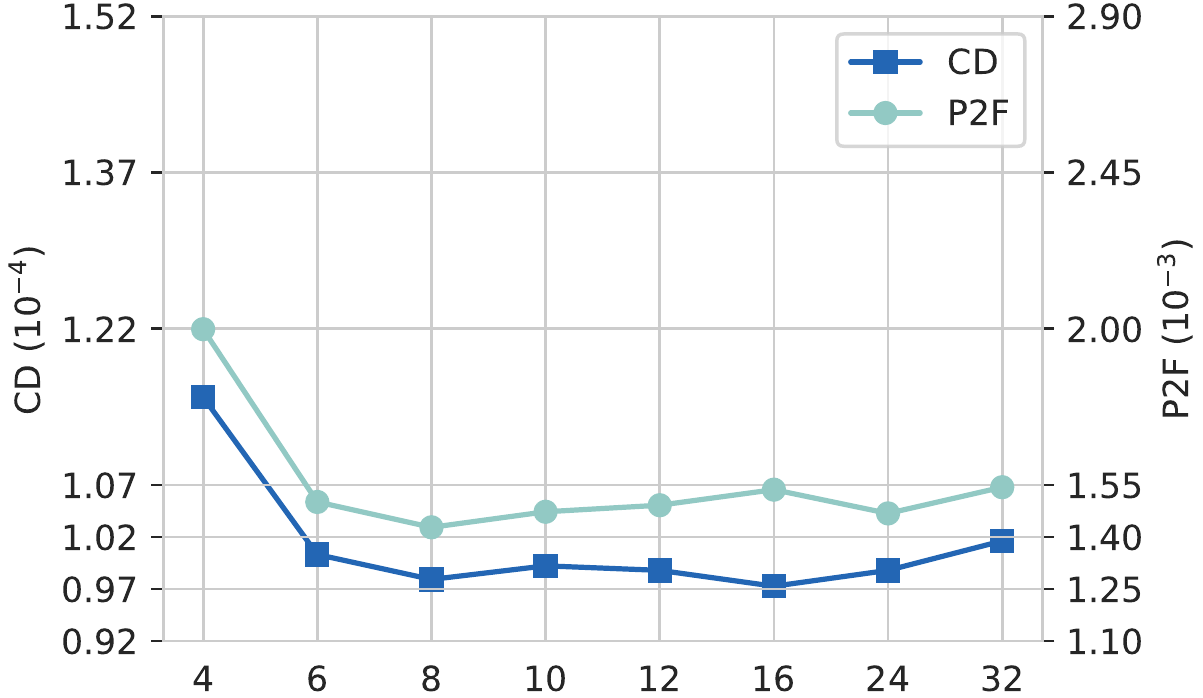}
\vspace{-0.1in}
\caption{Ablation study of number of interpolation neighbors for $K\in [4,32]$ to $K=32$.}
\label{fig:ablation-num-neighbors}
\end{figure}

\noindent
\textbf{Interpolation module.}
Fig. \ref{fig:ablation-num-neighbors} shows the impact of the number of neighbours $K$ participating in interpolation.
A larger value of $K$ means a broader area for generation.
We observe that both a small (\eg $K\leq4$) and large (\eg $K\geq32$) value of $K$ can lead to degraded performance.
Otherwise, our method is not sensitive to $K$ assignment.
In this study, we set $K=16$ by default.

Furthermore, we try to use the weights from $I_{\theta}$ to directly interpolate points in xyz coordinates (the xyz-interpolation row in Table \ref{tab:ablation-flow-architectures}).
It turns out that these weights are not feasible in Euclidean space, demonstrating that $I_{\theta}$ adaptively learns weights specific to latent point under prior distribution.

We also investigate the impact of space for point embedding and interpolation (\ie the graph type used in $E_{\theta}$ and $I_{\theta}$), as shown in Table \ref{tab:ablation-flow-architectures}.
Employing interpolation in latent space means that the $k$-nearest-neighbour graph is dynamically constructed (denoted as dynamic graph) in KNN operator of Fig. \ref{fig:illustration-network-arch}.
We can see a significant performance drop when using dynamic graph for interpolation.
We hypothesize the potential reason is as follows:
$F_{\theta}$ and $I_{\theta}$ are independent branch of upsampling pipeline.
Using dynamic graph in $I_{\theta}$ does not ensure consistent neighbour relationship between upsampled latents $\hat{z}^{R}$ and conditional features $c$, resulting into inconsistent features conditioning during inverse propagation.
In contrast, employing interpolation in Euclidean space (denoted as static graph) would not cause this issue, and thus achieve competitive performance.
Besides, the graph type used in $E_{\theta}$ has relative little impact on performance.

\begin{figure}[t]
\centering
\includegraphics[width=\linewidth, trim={0.05in, 0.15in, 0.05in, 0.05in}, clip]{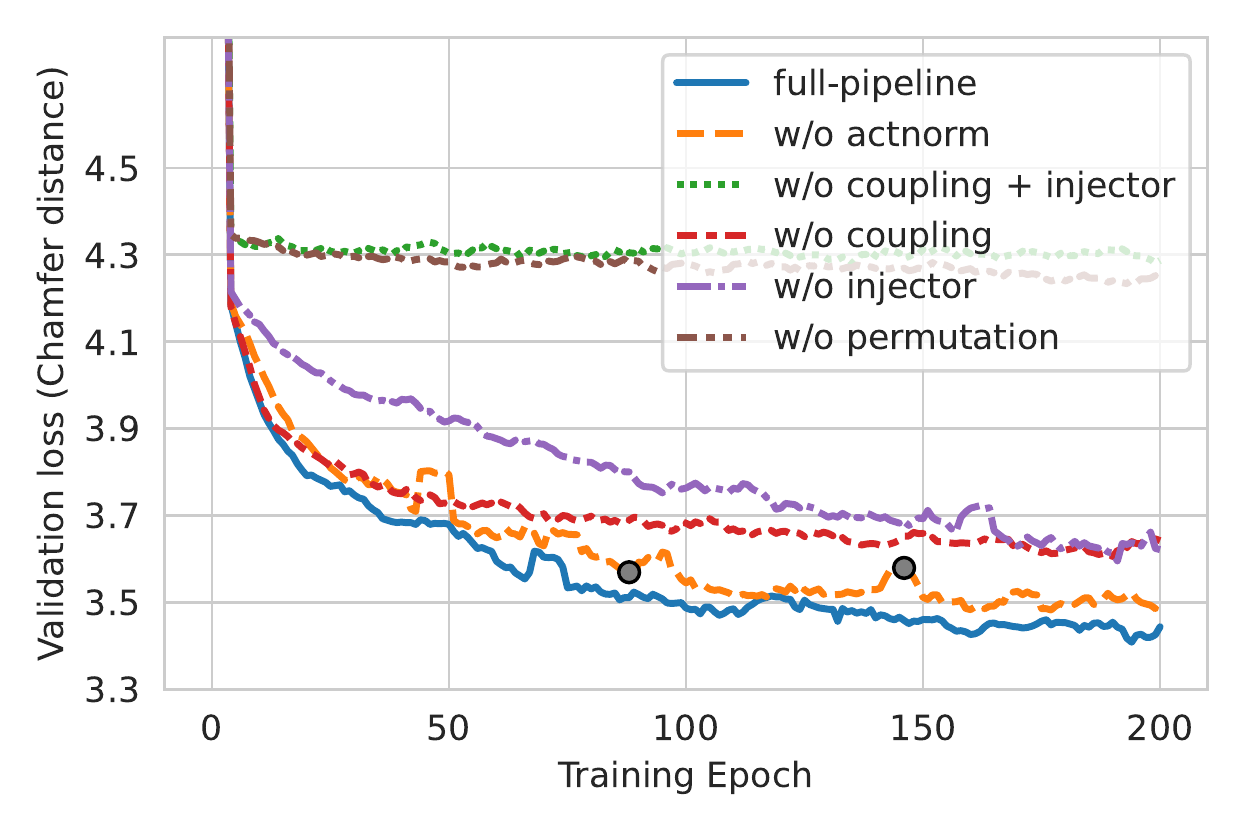}
\vspace{-0.25in}
\caption{
Ablation study of flow components.
We show the loss curves of training PU-Flow under different settings of flow components.
The grey point indicate the infinite value encountered during training.
}
\label{fig:ablation-flow-components}
\end{figure}

\noindent
\textbf{Flow components.}
Fig. \ref{fig:ablation-flow-components} shows the ablation study evaluated on each flow component.
We observe that our method cannot yield reasonable results without affine coupling/injector layers.
This demonstrates that the flow module $F_{\theta}$ requires features extracted from $E_{\theta}$ to encode point to latent representation.
The affine coupling and injector layers can be used as drop-in replacement to each other.
The permutation layers are necessary components to ensure that all channels are fully processed by coupling layer.
When the actnorm layer is removed, the training process becomes more sensitive to gradient explode and we may encounter invalid loss value (\eg infinite).
Thus, the actnorm layer makes significant contribution to training stability.

\section{Conclusion}

We presented a novel point cloud upsampling framework, called PU-Flow, that leverages NFs and weight prediction technique.
With the invertible capability of NFs, we can transform the point clouds between Euclidean space and latent distribution in a lossless manner.
We formulate upsampling as local ensemble of latent variables, with the interpolation weights adaptively learned from the local neighborhood context.
Given sparse point cloud as input, our method can produce a dense output that provides a reasonable prediction of the underlying surface and also contains fine details.
Quantitative evaluation demonstrates that PU-Flow outperforms existing state-of-the-art works in terms of quality and efficiency.


While we have demonstrated high quality results of PU-Flow, our approach is still subject to a few limitations that can be addressed in follow-up work. 
First, the dimension bottleneck problem, described in Section \ref{section:network_point_embedding}, limits the expressive ability of latent points in flow transformation.
Second, due to the invertible requirement, some commonly adopted network designs, such as skip connection, code assignment, up-down-up expansion pattern, \etc are not feasible for flow architecture.
Third, our method upsamples points at the \textit{patch} level, and thus has a limited ability in inferring global shape and large holes.

In the future, we will extend PU-Flow to simultaneously generate normals and a higher resolution of geometry for sparse input.
Furthermore, we will investigate the propagation pipeline of PU-Flow to point cloud compression tasks, which proposes a high requirement of detail reconstruction and efﬁcient storage, and denoise task, which is sensitive to noisy point distribution.


%

\ifCLASSOPTIONcaptionsoff
  \newpage
\fi



\bibliographystyle{IEEEtran}
\bibliography{egbib}

\vfill

\begin{IEEEbiography}[{\includegraphics[width=1in,height=1.25in,clip,keepaspectratio]{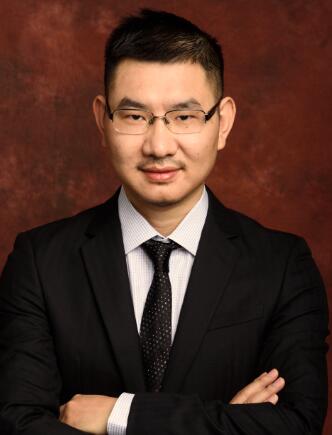}}]{Aihua Mao}
is a professor with the School of Computer Science and Engineering, South China University of Technology (SCUT), China.
He received the PhD degree from the Hong Kong Polytechnic University in 2009, the M.Sc degree from Sun Yat-Sen University in 2005 and the B.Eng degree from Hunan University in 2002.
His research interests include 3D vision and computer graphics.
\end{IEEEbiography}



\begin{IEEEbiography}[{\includegraphics[width=1in,height=1.25in,clip,keepaspectratio]{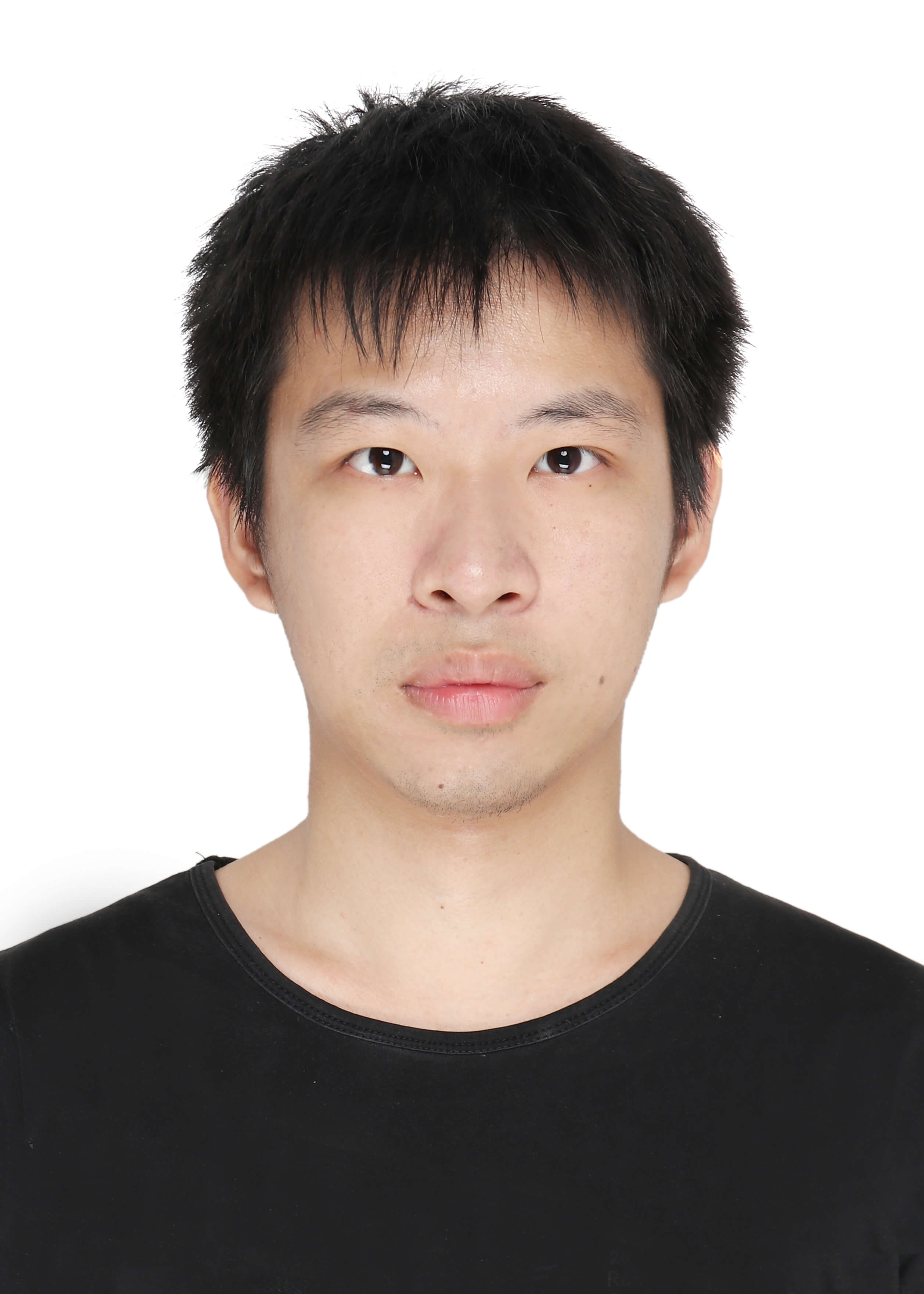}}]{Zihui Du}
received the B.S. degree in computer science from TianGong University, Tianjin, China in 2018.
He is currently pursuing the M.S. degree in computer science with South China University of Technology, Guangzhou, China.
His current research interests include machine learning, 3D point cloud generation and normalizing flows.
\end{IEEEbiography}


\begin{IEEEbiography}[{\includegraphics[width=1in,height=1.25in,clip,keepaspectratio]{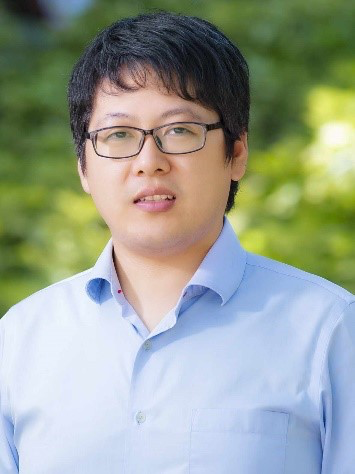}}]{Junhui Hou}
Junhui Hou received the B.Eng. degree in information engineering (Talented Students Program) from the South China University of Technology, Guangzhou, China, in 2009, the M.Eng. degree in signal and information processing from Northwestern Polytechnical University, Xian, China, in 2012, and the Ph.D. degree in electrical and electronic engineering from the School of Electrical and Electronic Engineering, Nanyang Technological University, Singapore, in 2016. Since Jan. 2017, he has been an Assistant Professor with the Department of Computer Science, City University of Hong Kong. His research interests fall into the general areas of visual computing, such as image/video/3D geometry data representation, processing and analysis, semi/un-supervised data modeling, and data compression and adaptive transmission.
\end{IEEEbiography}


\begin{IEEEbiography}[{\includegraphics[width=1in,height=1.25in,clip,keepaspectratio]{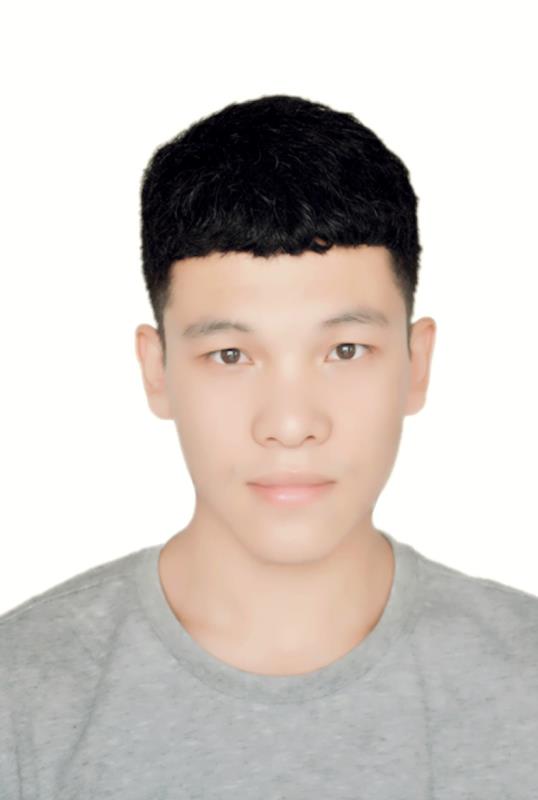}}]{Yaqi Duan}
received the B.S. degree in computer science from TianGong University, Tianjin, China in 2020. He is currently pursuing the M.S. degree in computer science with South China University of Technology, Guangzhou, China. His research focuses on deep learning, invertible networks and point cloud processing.
\end{IEEEbiography}


\begin{IEEEbiography}[{\includegraphics[width=1in,height=1.25in,clip,keepaspectratio]{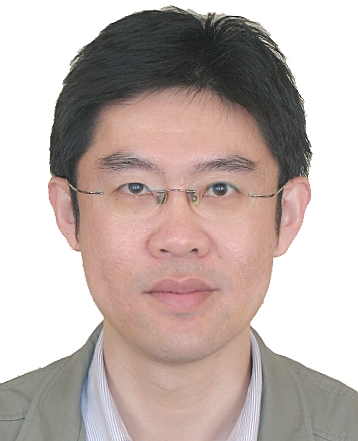}}]{Yong-Jin Liu}
is a professor with the Department of Computer Science and Technology, Tsinghua University, China.
He received the BEng degree from Tianjin University, China, in 1998, and the PhD degree from the Hong Kong University of Science and Technology, Hong Kong, China, in 2004.
His research interests include computer graphics and computer-aided design.
\end{IEEEbiography}


\begin{IEEEbiography}[{\includegraphics[width=1in,height=1.25in,clip,keepaspectratio]{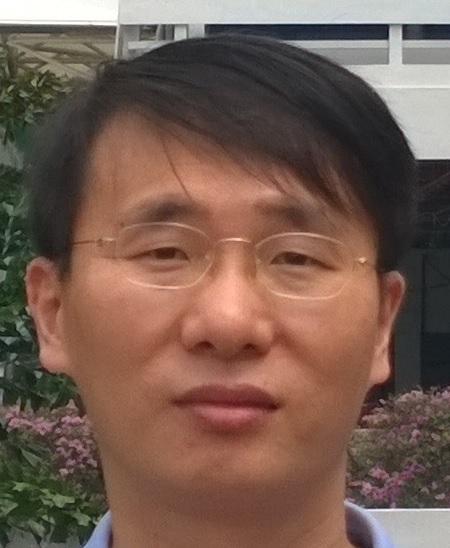}}]{Ying He}
is an associate professor in the School of Computer Engineering, Nanyang Technological University, Singapore.
He received his Bachelor (1997) and Master (2000) degrees in Electrical Engineering from Tsinghua University, and PhD (2006) in Computer Science from Stony Brook University.
He is interested in the problems that require geometric computing and analysis.
\end{IEEEbiography}

\vfill

\ifdefined\SHOWSUPPLEMENT

\clearpage
\newpage

\def\thesection{\Alph{section}}

\setcounter{section}{0}

\title{Supplementary Material for \\“PU-Flow: a Point Cloud Upsampling Network\\with Normalizing Flows”}
\supplementmaketitle

\section{Network Configurations}
\label{section:supplement-network-configuration}
In this section, we provide more details about the network architecture used in this paper.

\subsection{Point Embedding Unit}
\label{section:supplement-feat-extraction-unit}

In this study, we use a modified version of EdgeConv from dynamic graph CNN \cite{wang2019dynamic} as point embedding unit $E_{\theta}$.

\begin{figure}[h]
\centering
\includegraphics[width=\linewidth]{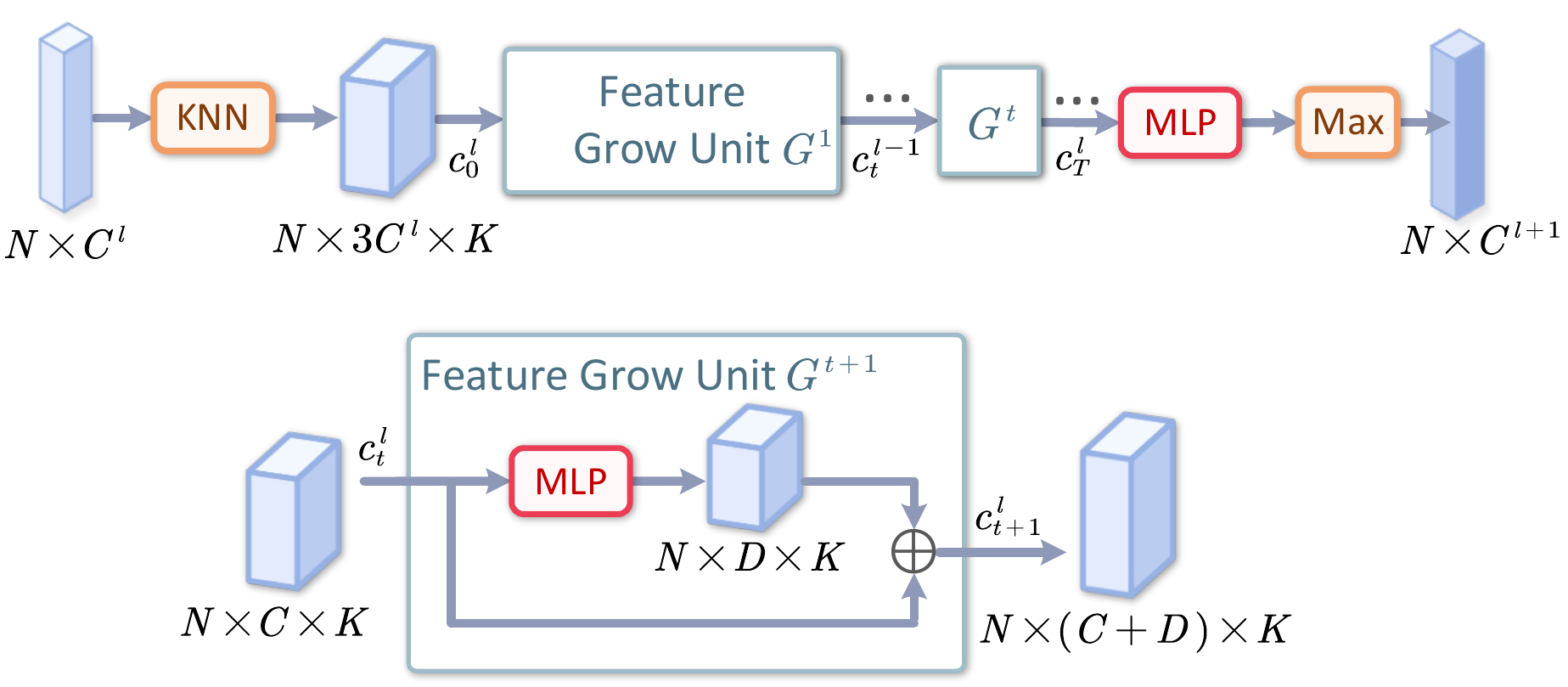}
\caption{Architecture of point embedding unit $E^l_{\theta}$.}
\label{fig:arch-point-embedding}
\end{figure}

To be specific, we first use $k$NN operator to build graph features as done in \cite{wang2019dynamic}.
Then, point features are sent to a sequence of Feature Grow Unit $E_G^t$ as shown in Fig. \ref{fig:arch-point-embedding}, where $T$ is the total number of $E_G^t$ units.
Finally, we employ a simple MLP and max operator to derive point-wise features as output of $E_{\theta}^l$.

In Feature Grow Unit $E_G^t$, we grow $D$ channels for each $E_G^t$ unit and then concatenate input features, where $D$ is generally a small number (\eg $D=8$).
In this way, we fully utilize features extracted from multi-stages including both low and high-level features.
Each MLP block consists of Conv2d, BatchNorm2d and ReLU layers.

\subsection{Discrete Flow Components}
\label{section:supplement-flow-components}
In this section, we briefly review flow layers along with the extended conditional setting \cite{ardizzone2019guided}, \cite{winkler2019learning}.
All flow components are required to be carefully designed to ensure invertibility and tractable Jacobian determinant \cite{dinh2014nice}, \cite{dinh2016density}.

\textbf{Conditional affine coupling.}
The affine coupling layer, proposed in RealNVP \cite{dinh2016density}, is a simple yet flexible invertible layer.
In the extension of conditional setting of affine coupling layer (Fig. \ref{fig:illustration-flow-block}), let $h^l\in \mathbb{R}^{N\times D}$ represent the output of the $l$-th flow layer.
Then, $ h^{l+1}$ follows the equation:
\begin{equation}\label{formula_affine_coupling_forward}
\left\{\begin{array}{l} h_{1: d}^{l+1}\ \ \ =\;h_{1: d}^l \\h_{d+1: D}^{l+1}=\;h_{d+1: D}^l \odot \exp \left(h_s^l\right) + h_b^l,\\\end{array}\right.
\end{equation}
where $h_s^l=f^l_{\theta,s} (h_{1: d}^l; c^l)$ and $h_b^l=f^l_{\theta,b}(h_{1: d}^l; c^l)$ are scale and bias terms of affine transform, $h^l=(h_{1: d}^l, h_{d+1: D}^l)$ is a partition of $h^l$ along the channel dimension, $\odot$ indicates element-wise product, and $d$ is partition location (where $d=D / 2$ in all experiment).
$c^l\in \mathbb{R}^{N\times C}$ represents conditional features derived from point embedding module $E_{\theta}$ (Section \ref{section:network_point_embedding}).
The inverse process of eq. (\ref{formula_affine_coupling_forward}) is:
\begin{equation}\label{formula_affine_coupling_inverse}
\left\{\begin{array}{l}h_{1: d}^l\;\;\;\ =\ \ h_{1: d}^{l+1} \\h_{d+1: D}^l = (h_{d+1: D}^{l+1} - h_b^l ) \odot \exp (-h_s^l).\end{array}\right.
\end{equation}
Note that the Jacobian matrix of eq. (\ref{formula_affine_coupling_forward}) is triangular, which means that its determinant can be simplified as product of diagonal elements.
Thus, the log-determinant of Jacobian can be calculated by summing up all channels of $h_s^l$, namely, $\log \left|\operatorname{det} \frac{\partial f_{\theta}^{l}}{\partial h^l}\right|=\sum_{i d} h_s^l$, where $i$ and $d$ are indexes of point and channel, respectively.

The transformation net $f^l_{\theta,s}(\cdot)$ and $f^l_{\theta,b}(\cdot)$ accept $c=[c^l,h^l]$ as input.
They can be arbitrarily complex neural networks from $\mathbb{R}^{d} \mapsto \mathbb{R}^{D-d}$ and are not required to be invertible.
In this study, $f_{\theta, s}^{l}(\mathbf{\cdot})$ and $f_{\theta, b}^{l}(\mathbf{\cdot})$ are both simply parameterized as fully connected layers.



\textbf{Permutation.}
To improve the nonlinear transformation capacity for flow layer, channels shuffle methods such as reverse and random permutation are commonly used.
Glow \cite{kingma2018glow} proposes to use invertible 1x1 convolution, namely inv1x1, as the channel permutation operator.
In practice, we observe improved performance switching from random permutation to inv1x1. 
This layer can be expressed as
\begin{equation}\label{formula_inv1x1}
    h^{l+1}=W h^{l},
\end{equation}
where $W$ is a matrix of size $D \times D$. The log-determinant of $W$ can be computed as $N \cdot \log |\det(W)|$, that is, $\log \left|\operatorname{det} \frac{\partial f_{\theta}^{l}}{\partial h^l}\right|=N \cdot \log |\det(W)|$. The inverse process of eq. (\ref{formula_inv1x1}) is $h^{l}=W^{-1} h^{l+1}$.

\textbf{Actnorm.}
The activation normalization layer, namely actnorm \cite{kingma2018glow}, performs channel-wise normalization with learnable scale $\mathbf \mu\in \mathbb{R}^{D}$ and bias $\mathbf \sigma\in \mathbb{R}^{D}$ to improve training stability as follows:
\begin{equation}\label{formula_actnorm}
    h^{l+1}=\frac{h^{l}-\mathbf\mu}{\mathbf\sigma}.
\end{equation}
One can consider actnorm as a linear rescaling layer on each channel.
Its log-determinant can be computed as $\log \left|\operatorname{det} \frac{\partial f_{\theta}^{l}}{\partial h^l}\right|=N\cdot\sum_{d}{\mathbf\sigma}_d$.
The inverse process of eq. (\ref{formula_actnorm}) is denoted as $h^{l}=\mathbf\sigma h^{l+1}+\mathbf\mu$.

\textbf{Conditional Affine Injector.}
Inspired by SRFlow \cite{lugmayr2020srflow}, we introduce the affine injector layer to enhance feature conditioning.
Compared with the affine coupling layer, the injector layer generates scale and bias terms only depending on conditioning features $c^l$ as:
\begin{equation}\label{formula_affine_injector}
    h^{l+1}=\exp \left(f_{\theta, s}^{l}(c^l)\right) \cdot h^l+f^l_{\theta, b}(c^l).
\end{equation}
Similar to the coupling layer, the log-determinant of Jacobian is calculated by summing up all channels of $f_{\theta, \mathrm s}^l(\cdot)$.

\subsection{Continuous Flow Block}
Our continuous model consists of $L$ flow blocks.
Each continuous flow block is basically an ODE-Net implemented in FFJORD \cite{grathwohl2018ffjord}.
We follow the PointFlow \cite{yang2019pointflow}'s released code and use two \texttt{concatsquash} layers (with a hidden dimension of $64$) to model the continuous transformation $f_\theta$ and \texttt{tanh} as non-linear activation layer.
We do not use the Moving Batch Normalization layer in our implementation, as we find the model become unable to converge.

\subsection{Point Feature Extractor \texorpdfstring{$I_E$}{Lg}}

In Section \ref{section:network_interpolation}, we use feature extractor $I_E$ to learn analyze point-wise local geometric context $s_i\in\mathbb{R}^C$.
Technically, $I_E$ can be realized as any modern point-wise feature extractor.
In this study, $I_E$ can be formulated as follows:
\begin{equation}\label{formula:interp-feat-extractor}
s_i=[I_{E_1}(\mathcal P_i^K), I_{E_2}(\mathcal P_i^K)],
\end{equation}
where $\mathcal P_i^K\in \mathbb{R}^{K\times D}$ represents the set of $k$-nearest neighbors of point $p_i$.
$I_{E_1}$ and $I_{E_2}$ are EdgeConv unit \cite{wang2019dynamic} and Point Embedding unit (Section \ref{section:supplement-feat-extraction-unit}), respectively.
The $[\cdots]$ operator denotes concatenation operator along channel dimension.
We use static graph in both $I_{E_1}$ and $I_{E_2}$.


\subsection{Architecture of vanilla model}
In Section \ref{sec:ablation-study}, we compare the vanilla weighted interpolation model with our method.
As shown in Fig. \ref{fig:illustration-vanilla-arch}, this vanilla model realizes the upsampling pipeline with two basic branches: one is for weight estimation and another is for neighbour context extraction.
Basically, we reuse the feature extractor $I_E$ and weight estimator $I_W$ used in our interpolation module $I_\theta$.
We add another feature extractor $I_{F}$ to extract features on point-wise local neighbour.
Basically, $I_{F}$ is constructed by a sequence of Conv2D, BatchNorm2d, LeakyReLU layers.
The interpolation process is identical to Section \ref{section:network_interpolation}.
Finally, we use MLPs for coordinate reconstruction.

\section{Implementation Details}
\label{section:supplement-impl-details}

\subsection{Dataset Configurations}

\textbf{Training phase.}
Training point clouds are extracted from mesh models by Poisson disk sampling algorithm \cite{corsini2012efficient}, with 5K points for sparse models.
An input patch $\mathcal P$ of $N=256$ points and a ground-truth patch $\mathcal X$ of the corresponding upsampling factor are cropped in advance for training.
To increase data diversity and avoid overfitting, we adopted various data augmentation techniques to training patches, including point perturbation and random rotation.

\noindent
\textbf{Evaluation phase.}
We first normalized meshes into range $[-1, 1]$, and then used Poisson-disk sampling \cite{corsini2012efficient} to sample points input sparse points and ground-truth dense points.


During the evaluation phase, the point set is split into patches for independent upsampling.
Similar to the previous work, we first selected seed points from the input point set $\mathcal P$ by the farthest point sampling (FPS).
For each seed point, we extracted a local patch of 256 points by $k$NN with overlapped areas between patches.
Then, we fed the patches into PU-Flow for upsampling as shown in Fig. \ref{fig:illustration-network-arch}.
Thereafter, upsampled patches are merged, and finally are downsampled to the target number of points by FPS as the final result.

For the EAR algorithm, we use the code implementation from CGAL library \cite{cgal:eb-21b}.
All methods evaluated in Table \ref{tab:quantitative_pu1k} are retrained on PU1K training set.
\yh{
Note that, since PUGeo-Net\cite{qian2020pugeo} and MAFU\cite{QianFlexiblePU2021} both require normal data during training, which are not available in PU1K and PU-GAN dataset, we removed their normal generation module and then retrained them.
}
All methods evaluated in Table \ref{tab:quantitative_pugeo} are retrained on the PUGeo-Net training set.
\yh{
All methods evaluated in Table \ref{tab:quantitative_pugan} are retrained on the PU-GAN training set.
}
Table \ref{tab:quantitative_famousthingi} is evaluated on the same models as Table \ref{tab:quantitative_pugeo}.

\subsection{Network Training}
\label{section:supplement-network-training}
Basically, we use the same training settings for our discrete and continuous model.
We train our method with Adam optimizer, learning rate 0.001, and batch size 32.
For PU1K dataset, we train for 50 epoch, with 2216 batch each epoch.
For PUGeo-Net dataset, we train for 200 epoch, with 270 batch each epoch.
The latent distribution $z \sim p_{\vartheta}(z)$ is initialized as standard Gaussian with mean $0$ and variance $1$.
We use the checkpoint with minimum $\mathcal{L}_{\text{rec}}$ on the validation set as our final model.
The loss tuning hyper-parameters $\alpha$ and $\beta$ are empirically set as 1e\textsuperscript{-4}, and 5e\textsuperscript{-2}, respectively.

\section{More Experiments}
To further demonstrate the robustness of PU-Flow, we also evaluate it with noisy, sparse, and non-uniform data.
We collect some 3D printing models from Thingi10k \cite{zhou2016thingi10k} dataset for evaluation.

\textbf{Upsampling point sets of various noise levels.}
Fig. \ref{fig:more-expriment-noise} shows the input points under different noise levels and corresponding upsampled results by PU-Flow.
The noisy point clouds are generated by adding Gaussian noise with standard deviation increasing from 0.0\% to 2.0\% of bounding box diagonal.
Although the upsampled points become more unstable as the noise level increases, our method can still produce a consistent shape with the noise-free reference, indicating its robustness against various noise levels.

\textbf{Upsampling point sets of various input sizes.}
To confirm the robustness under sparsity, we upsample the input points of varying sizes (\eg numbers of points).
As shown in Fig. \ref{fig:more-expriment-size}, we observe that our method can generate absent high-frequency details on both sparse and dense inputs.

\textbf{Upsampling point sets of non-uniform data.}
Moreover, we show the upsampled result by PU-Flow with the non-uniform input in Fig. \ref{fig:more-expriment-non-uniform}.
We randomly sample 5,000 points from ground truth to simulate non-uniformly distributed points.
We observe that non-uniformity causes difficulty in preserving unified density across different areas, resulting in more holes and point clusters.
Compared with other methods, our method preserves better uniformity and produce less holes, validating its robustness against non-uniformity.


\section{Additional Qualitative Results}

In this section, we provide additional qualitative results as follows:

Fig. \ref{fig:visual-sota-more-points} shows full visual comparisons of upsampled points between various state-of-the-art methods.

Fig. \ref{fig:visual-sota-mesh-2} shows more visual comparisons of reconstructed mesh of various methods.

Fig. \ref{fig:more-visual-comparison-KITTI} shows more visual comparison on KITTI \cite{Geiger2013IJRR} dataset.

Fig. \ref{fig:scanobjectnn-part2-v1} shows more upsampled results on the ScanObjectNN \cite{uy-scanobjectnn-iccv19} dataset.


Fig. \ref{fig:supplement-dataset-models} shows all the collected mesh models in the PU36 dataset.

\begin{figure*}
\centering
\ifdefined\GENERALCOMPILE
    \ifdefined\HIGHRESOLUTIONFIGURE
        \includegraphics[width=0.5\linewidth]{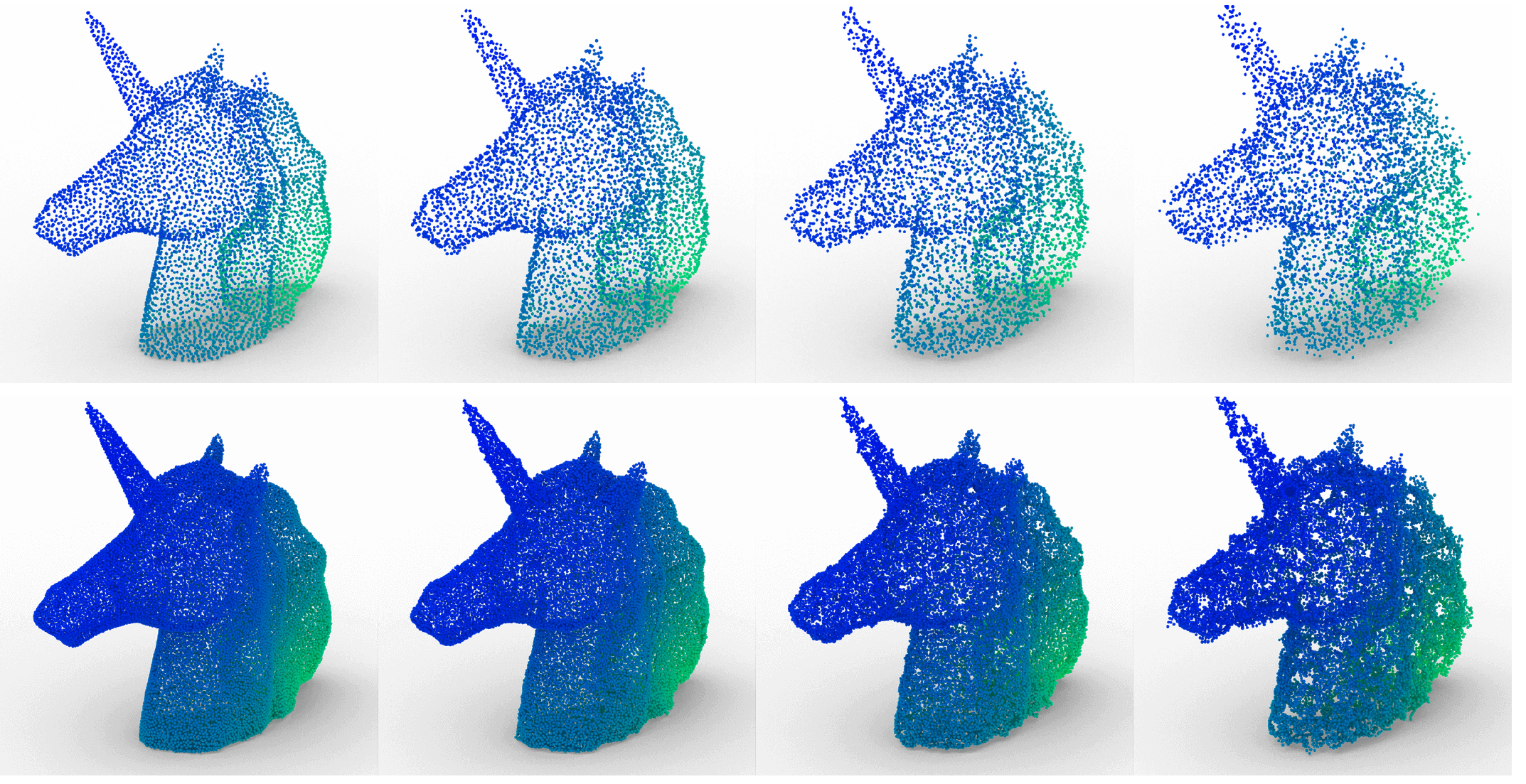}
    \else
        \includegraphics[width=0.5\linewidth]{images/plotting/reduce/other-noise-1.pdf}
    \fi
\fi
\hspace*{-.1in}
\hfill\vline\hfill
\ifdefined\GENERALCOMPILE
    \ifdefined\HIGHRESOLUTIONFIGURE
        \includegraphics[width=0.5\linewidth]{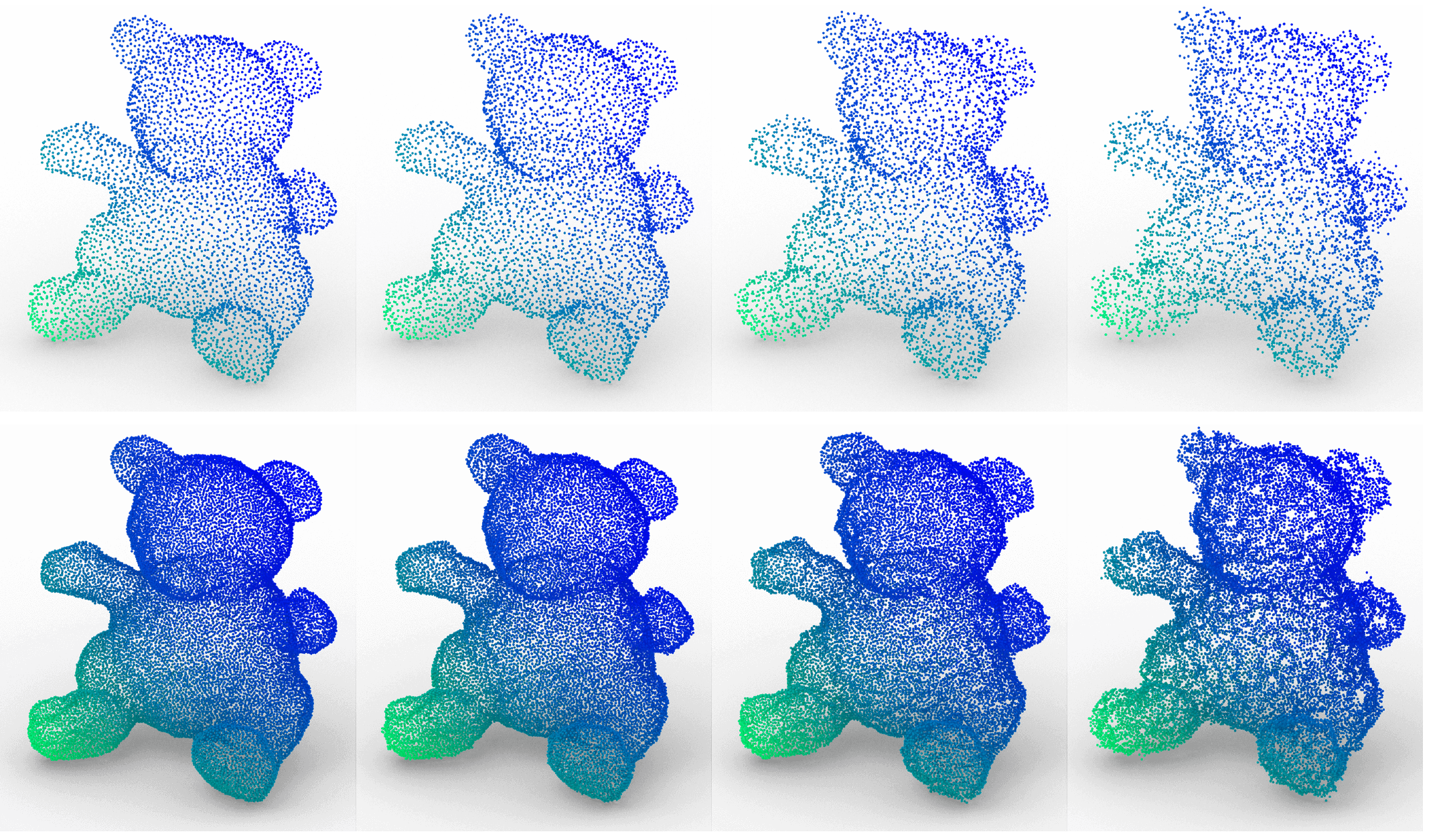}
    \else
        \includegraphics[width=0.5\linewidth]{images/plotting/reduce/other-noise-2.pdf}
    \fi
\fi

\begin{tabular}{@{}C{0.125\linewidth}@{}C{0.125\linewidth}@{}C{0.125\linewidth}@{}C{0.125\linewidth}@{}C{0.125\linewidth}@{}C{0.125\linewidth}@{}C{0.125\linewidth}@{}C{0.125\linewidth}@{}}
0.0\% & 0.5\% & 1.0\% & 2.0\% & 0.0\% & 0.5\% & 1.0\% & 2.0\% \\
\end{tabular}
\caption{Upsampling results (2rd row) from inputs (1st row) under various noisy level ($R=4$).}
\label{fig:more-expriment-noise}
\end{figure*}

\begin{figure*}
\centering
\ifdefined\GENERALCOMPILE
    \ifdefined\HIGHRESOLUTIONFIGURE
        \includegraphics[width=0.5\linewidth]{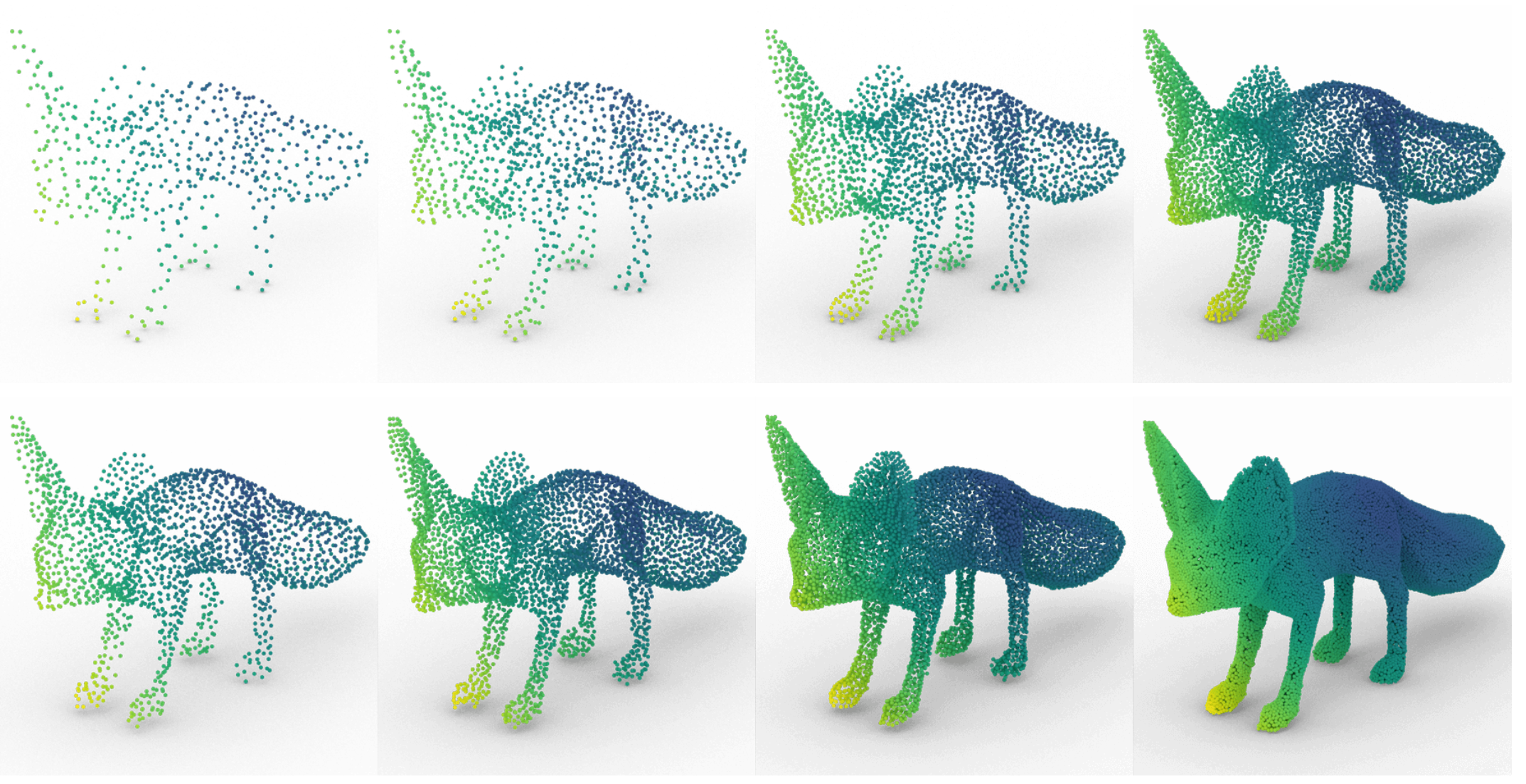}
    \else
        \includegraphics[width=0.5\linewidth]{images/plotting/reduce/other-size-1.pdf}
    \fi
\fi
\hspace*{-.1in}
\hfill\vline\hfill
\ifdefined\GENERALCOMPILE
    \ifdefined\HIGHRESOLUTIONFIGURE
        \includegraphics[width=0.5\linewidth]{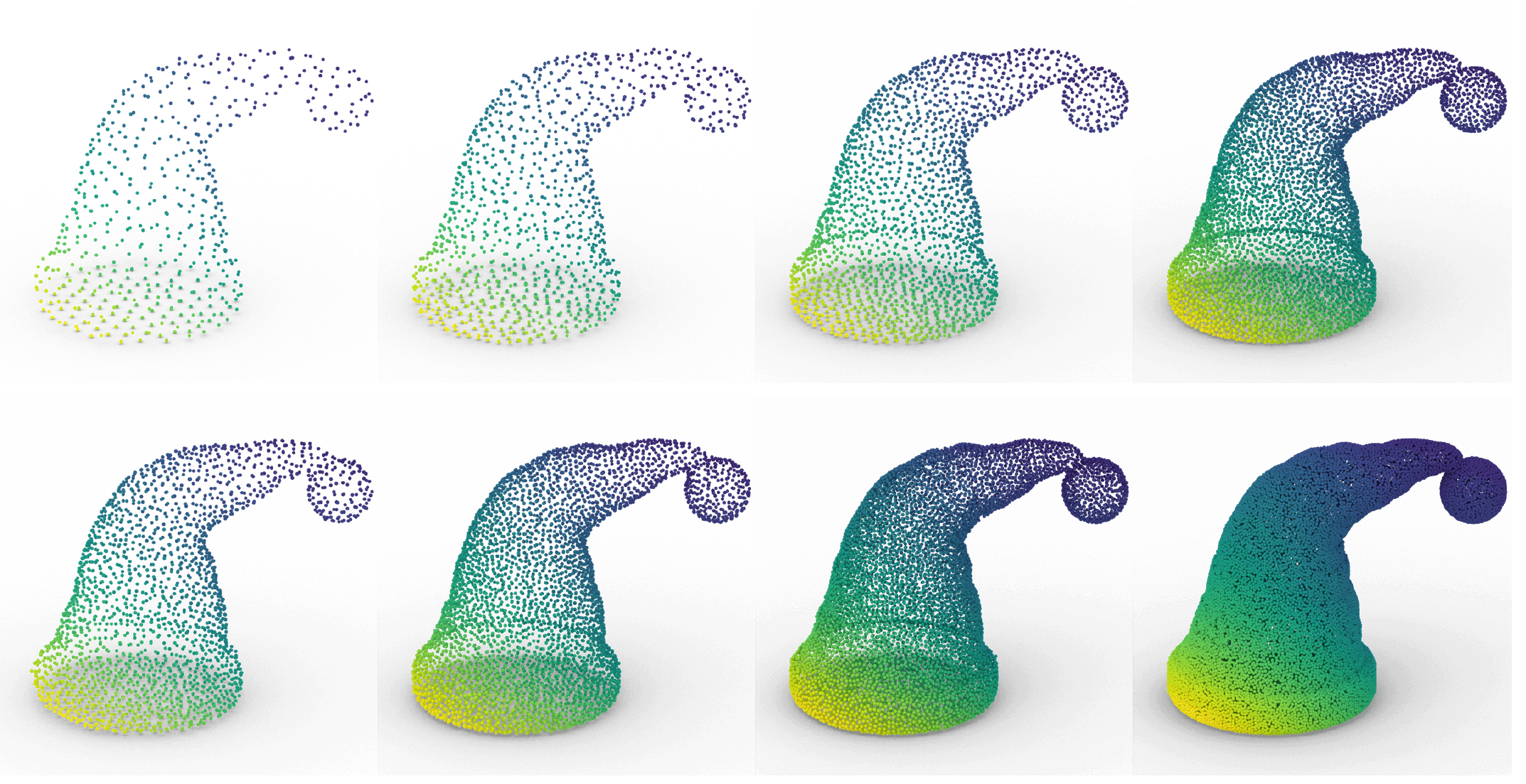}
    \else
        \includegraphics[width=0.5\linewidth]{images/plotting/reduce/other-size-2.pdf}
    \fi
\fi

\begin{tabular}{@{}C{0.125\linewidth}@{}C{0.125\linewidth}@{}C{0.125\linewidth}@{}C{0.125\linewidth}@{}C{0.125\linewidth}@{}C{0.125\linewidth}@{}C{0.125\linewidth}@{}C{0.125\linewidth}@{}}
$N=512$ & $N=1024$ & $N=2048$ & $N=5000$ & $N=512$ & $N=1024$ & $N=2048$ & $N=5000$ \\
\end{tabular}
\caption{Upsampling results from inputs of varying sizes ($R=4$).}
\label{fig:more-expriment-size}
\end{figure*}

\begin{figure*}
\centering
\hspace*{-.1in}
\ifdefined\GENERALCOMPILE
    \ifdefined\HIGHRESOLUTIONFIGURE
        \includegraphics[width=\linewidth]{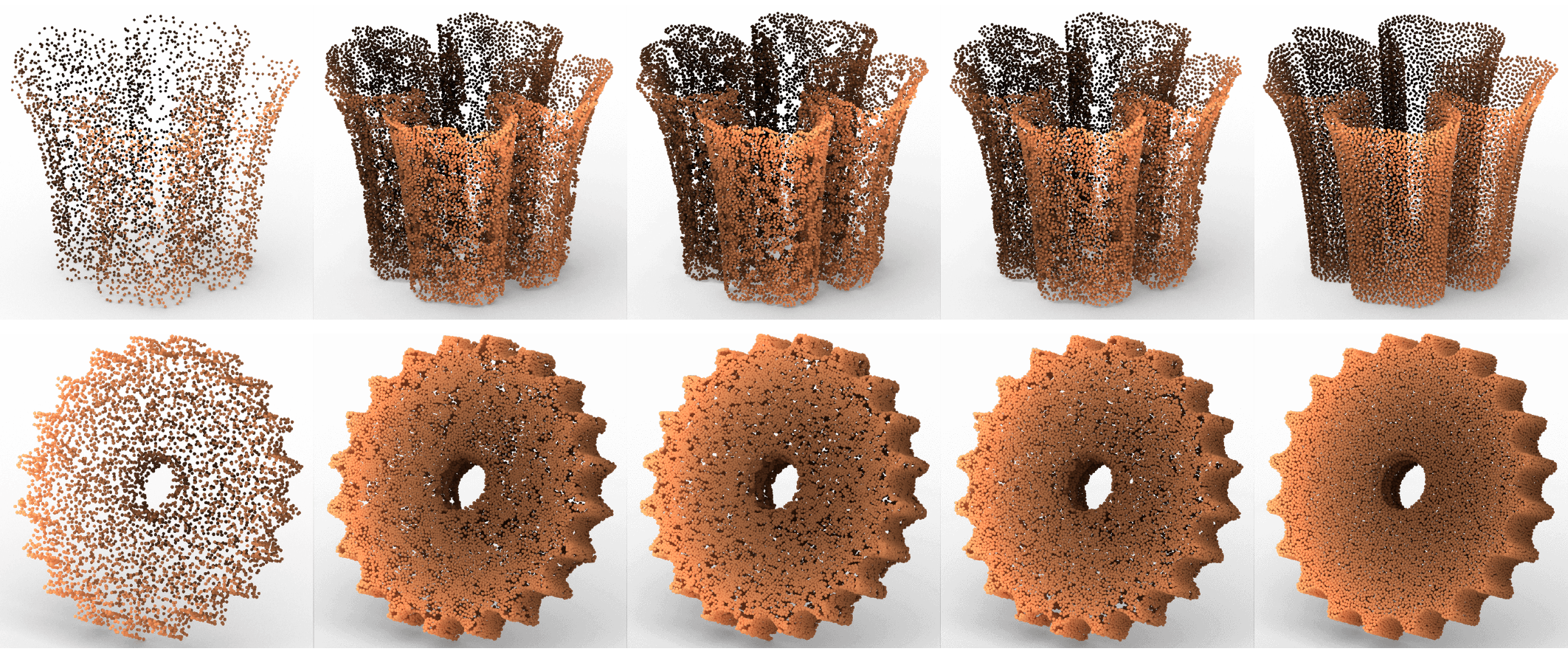}
    \else
        \includegraphics[width=\linewidth]{images/plotting/reduce/other-non-uniform.pdf}
    \fi
\fi
\begin{tabular}{@{}C{0.20\linewidth}@{}C{0.20\linewidth}@{}C{0.20\linewidth}@{}C{0.20\linewidth}@{}C{0.20\linewidth}@{}}
Input & Dis-PU \cite{li2021dispu} & PUGeo-Net \cite{qian2020pugeo} & Ours & Ground Truth \\
\end{tabular}
\caption{Upsampling results from non-uniform input ($R=4$).}
\label{fig:more-expriment-non-uniform}
\end{figure*}

\begin{sidewaysfigure*}
\centering
\hspace*{\fill}  
\begin{tikzpicture}
\node (fig) at (current page.east) {\includegraphics[width=0.15\textwidth]{images/plotting/colorbar-points.pdf}};
\node[left=0cm of fig] {Low};
\node[right=0cm of fig] {High};
\end{tikzpicture}

\ifdefined\GENERALCOMPILE
    \ifdefined\HIGHRESOLUTIONFIGURE
        \includegraphics[width=\linewidth]{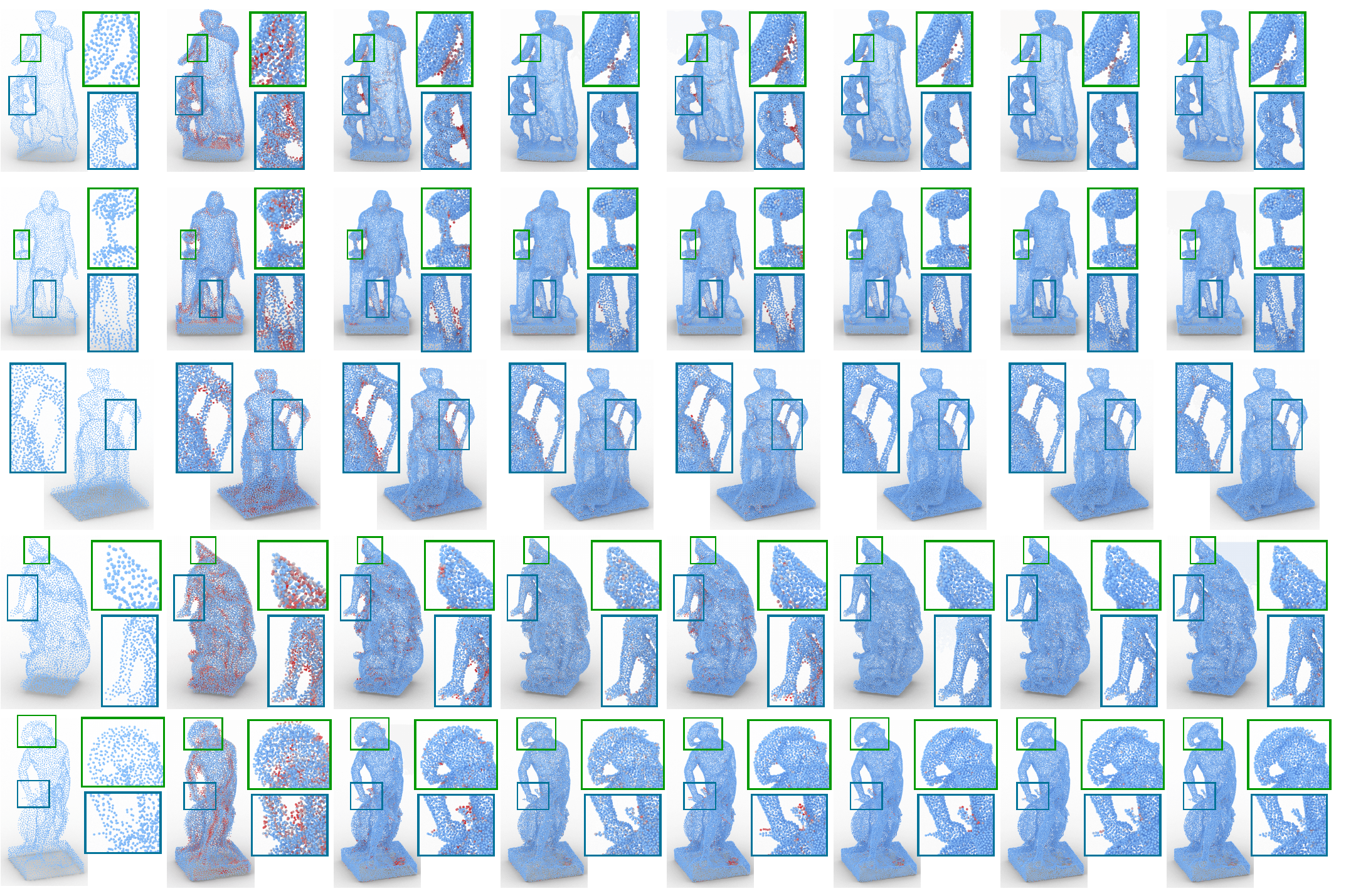}
    \else
        \includegraphics[width=\linewidth]{images/plotting/reduce/sota-more-points-visual-comparison-1-reduce.pdf}
    \fi
\fi

\begin{tabularx}{\linewidth}{@{}Y@{}Y@{}Y@{}Y@{}Y@{}Y@{}Y@{}Y@{}}
(a) Input &
(b) PU-Net \cite{Yu2018PUNetPC} &
(c) PU-GAN\cite{li2019pugan} &
(d) PU-GCN \cite{Qian_2021_CVPR} &
(e) Dis-PU \cite{li2021dispu} &
(f) MAFU \cite{QianFlexiblePU2021} &
(g) PUGeo-Net \cite{qian2020pugeo} &
(h) Ours (discrete) \\
\end{tabularx}
\caption{
More visual comparisons of various methods (b-h).
We visualized the P2F errors by colors for each point.
}
\end{sidewaysfigure*}

\begin{sidewaysfigure*}
\ContinuedFloat
\centering
\hspace*{\fill}  
\begin{tikzpicture}
\node (fig) at (current page.east) {\includegraphics[width=0.15\textwidth]{images/plotting/colorbar-points.pdf}};
\node[left=0cm of fig] {Low};
\node[right=0cm of fig] {High};
\end{tikzpicture}

\ifdefined\GENERALCOMPILE
    \ifdefined\HIGHRESOLUTIONFIGURE
        \includegraphics[width=\linewidth]{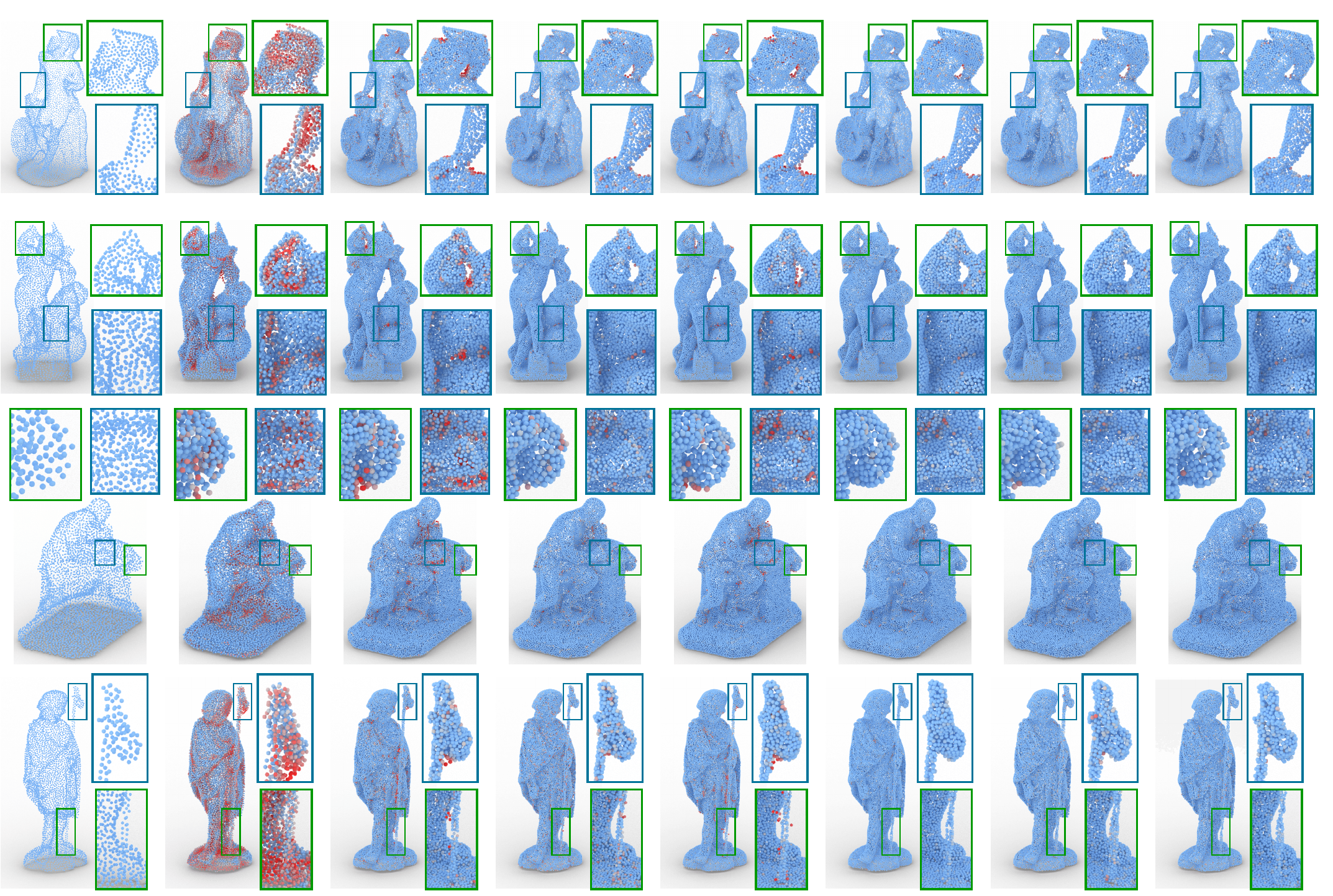}
    \else
        \includegraphics[width=\linewidth]{images/plotting/reduce/sota-more-points-visual-comparison-2-reduce.pdf}
    \fi
\fi

\begin{tabularx}{\linewidth}{@{}Y@{}Y@{}Y@{}Y@{}Y@{}Y@{}Y@{}Y@{}}
(a) Input &
(b) PU-Net \cite{Yu2018PUNetPC} &
(c) PU-GAN\cite{li2019pugan} &
(d) PU-GCN \cite{Qian_2021_CVPR} &
(e) Dis-PU \cite{li2021dispu} &
(f) MAFU \cite{QianFlexiblePU2021} &
(g) PUGeo-Net \cite{qian2020pugeo} &
(h) Ours (discrete) \\
\end{tabularx}
\caption{
More visual comparisons of various methods (b-h).
We visualized the P2F errors by colors for each point.
}
\label{fig:visual-sota-more-points}
\end{sidewaysfigure*}


\begin{figure*}[t]
\centering
\ifdefined\GENERALCOMPILE
    \ifdefined\HIGHRESOLUTIONFIGURE
        \includegraphics[width=\linewidth]{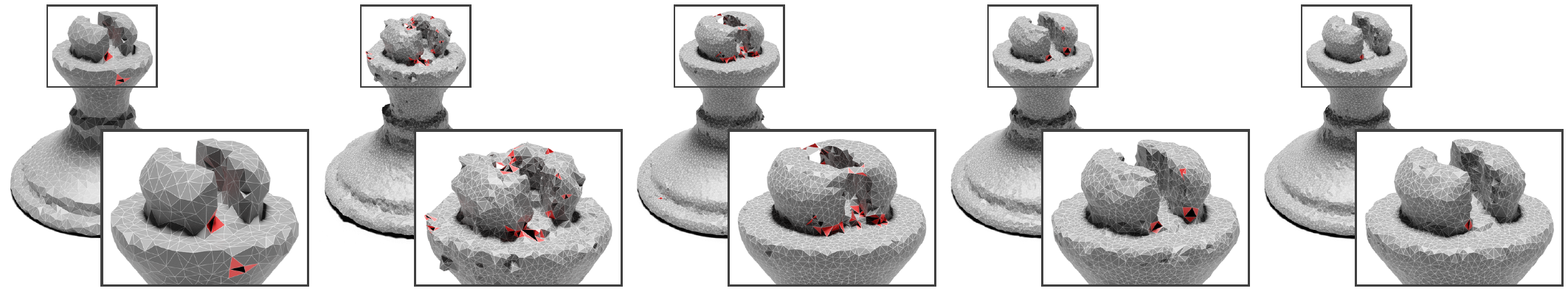}
    \else
        \includegraphics[width=\linewidth]{images/plotting/reduce/ChairBase.pdf}
    \fi
\fi
\begin{tabularx}{\linewidth}{@{}Y@{}Y@{}Y@{}Y@{}Y@{}}
0.348/0.16\%/7.0 &
0.383/1.02\%/8.2 &
0.354/0.75\%/7.5 &
0.337/0.05\%/7.3 &
0.333/0.05\%/7.1 \\
\end{tabularx}
\ifdefined\GENERALCOMPILE
    \ifdefined\HIGHRESOLUTIONFIGURE
        \includegraphics[width=\linewidth]{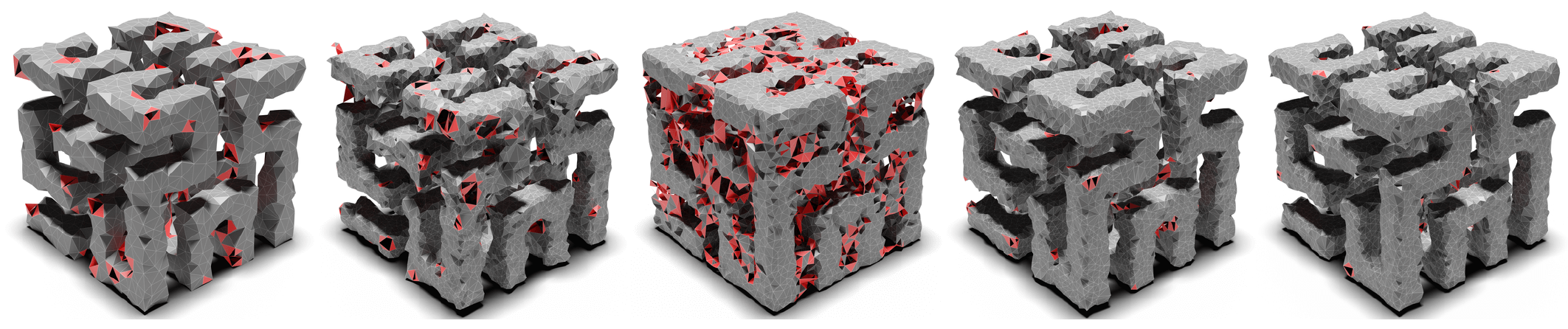}
    \else
        \includegraphics[width=\linewidth]{images/plotting/reduce/hilbert_cube1.pdf}
    \fi
\fi
\begin{tabularx}{\linewidth}{@{}Y@{}Y@{}Y@{}Y@{}Y@{}}
0.817/5.48\%/17.9 &
0.930/1.76\%/17.5 &
1.488/16.9\%/35.0 &
0.734/1.59\%/16.5 &
0.745/1.03\%/16.7 \\
\end{tabularx}
\ifdefined\GENERALCOMPILE
    \ifdefined\HIGHRESOLUTIONFIGURE
        \includegraphics[width=\linewidth]{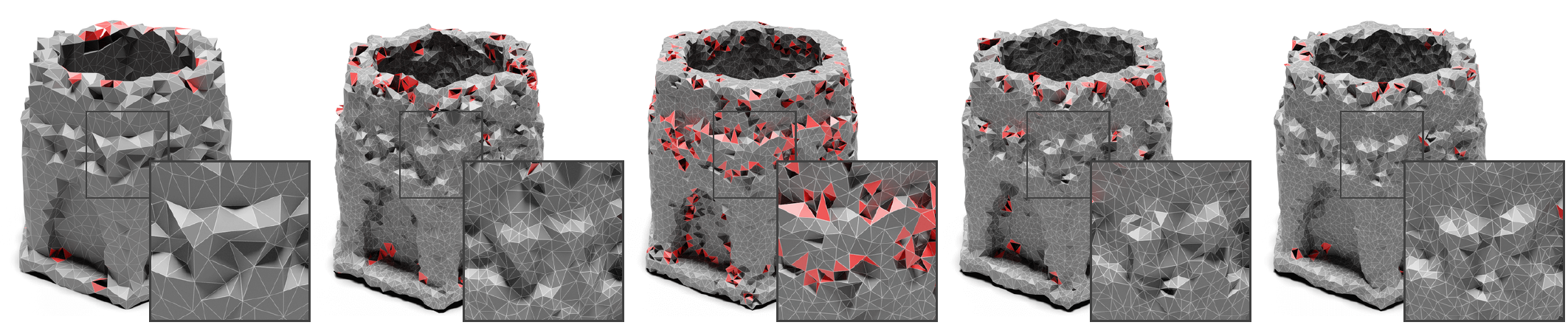}
    \else
        \includegraphics[width=\linewidth]{images/plotting/reduce/caer1.pdf}
    \fi
\fi
\begin{tabularx}{\linewidth}{@{}Y@{}Y@{}Y@{}Y@{}Y@{}}
0.617/1.57\%/13.5 &
0.734/1.92\%/16.6 &
0.869/12.0\%/24.6 &
0.612/1.72\%/14.4 &
0.621/1.66\%/15.5 \\
\end{tabularx}
\ifdefined\GENERALCOMPILE
    \ifdefined\HIGHRESOLUTIONFIGURE
        \includegraphics[width=\linewidth]{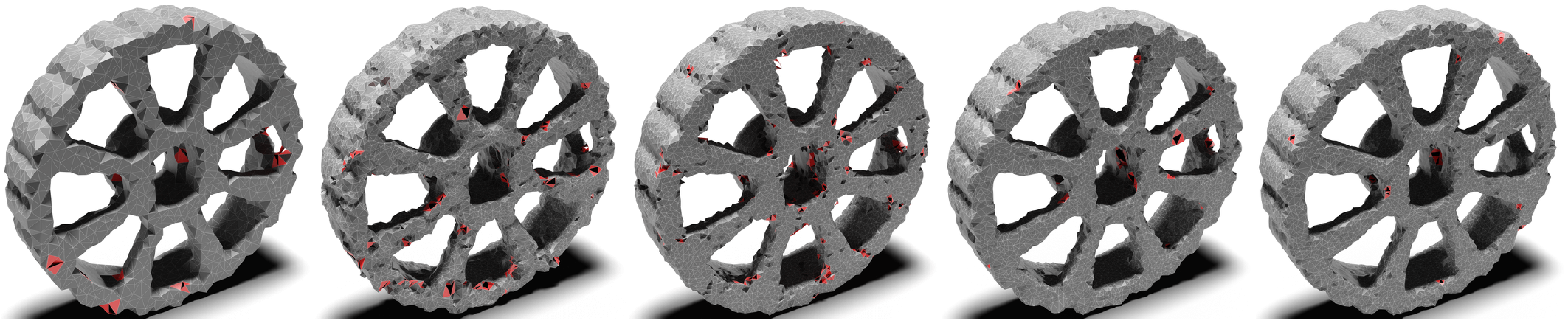}
    \else
        \includegraphics[width=\linewidth]{images/plotting/reduce/Handle.pdf}
    \fi
\fi
\begin{tabularx}{\linewidth}{@{}Y@{}Y@{}Y@{}Y@{}Y@{}}
0.468/1.53\%/13.4 &
0.504/3.27\%/16.2 &
0.478/3.02\%/15.0 &
0.434/0.72\%/12.1 &
0.429/0.35\%/11.9 \\
\end{tabularx}
\ifdefined\GENERALCOMPILE
    \ifdefined\HIGHRESOLUTIONFIGURE
        \includegraphics[width=\linewidth]{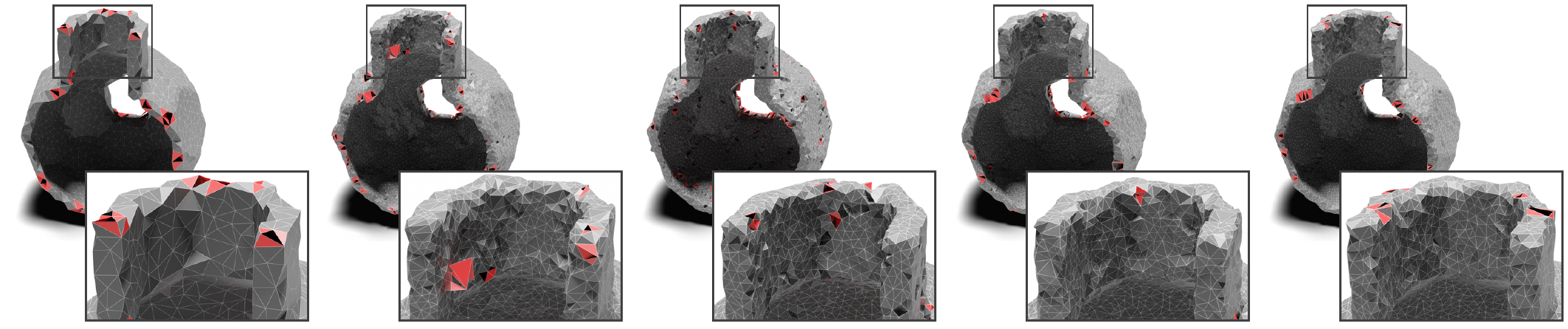}
    \else
        \includegraphics[width=\linewidth]{images/plotting/reduce/1.pdf}
    \fi
\fi
\begin{tabularx}{\linewidth}{@{}Y@{}Y@{}Y@{}Y@{}Y@{}}
0.407/1.87\%/6.84 &
0.474/1.21\%/10.9 &
0.488/2.51\%/12.6 &
0.415/0.83\%/8.16 &
0.410/0.63\%/7.19 \\
\end{tabularx}
\vskip 0.05in

\begin{tabularx}{\linewidth}{@{}Y@{}Y@{}Y@{}Y@{}Y@{}}
(a) Input &
(b) PU-GCN \cite{Qian_2021_CVPR} &
(c) Dis-PU \cite{li2021dispu} &
(d) PUGeo-Net \cite{qian2020pugeo} &
(e) Ours (discrete) \\
\end{tabularx}

\vspace*{-0.1in}
\caption{More visual comparisons of reconstructed surfaces of various methods (b-e).}
\label{fig:visual-sota-mesh-2}
\end{figure*}

\begin{figure*}[t]
\centering
\includegraphics[width=\textwidth]{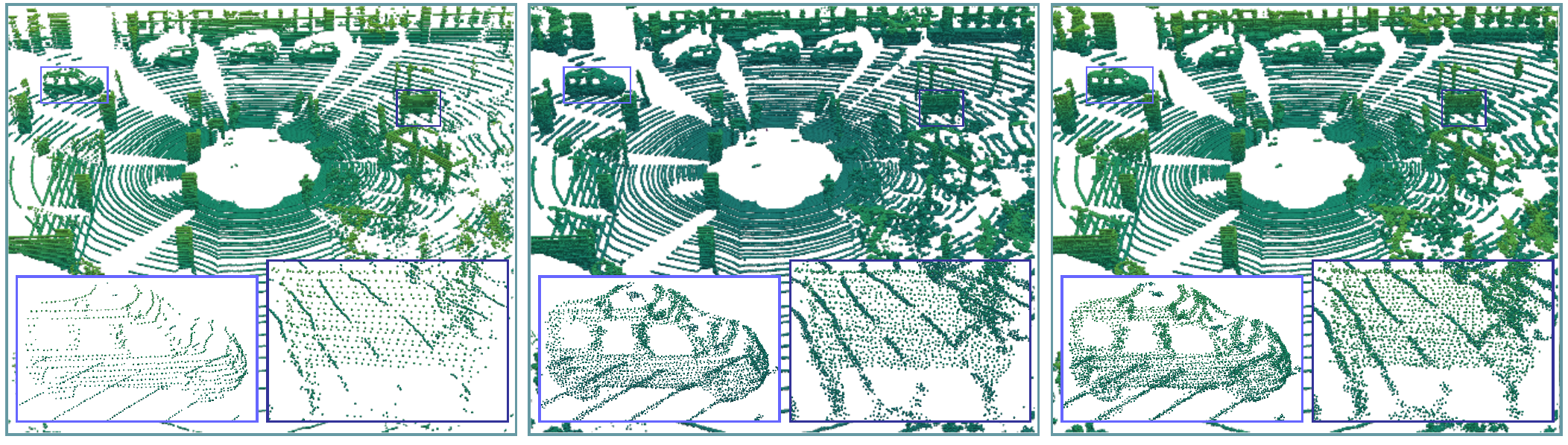}
\begin{tabularx}{\textwidth}{@{}Y@{}Y@{}Y@{}}
(a) Input & (b) PU-GCN \cite{Qian_2021_CVPR} & (c) PUGeo-Net \cite{qian2020pugeo}
\end{tabularx}
\includegraphics[width=\textwidth]{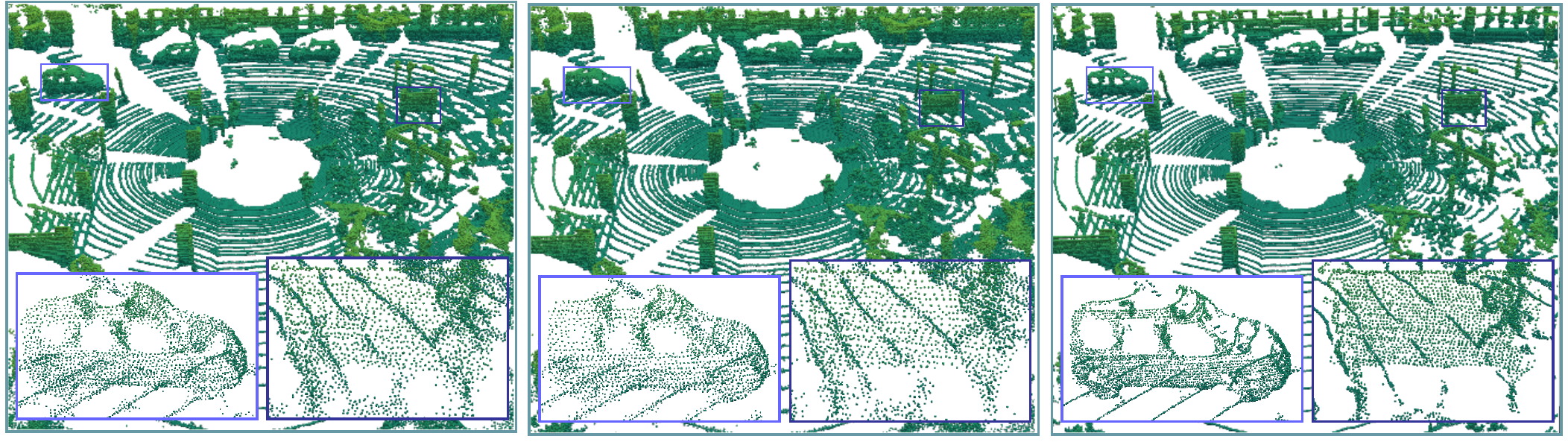}
\begin{tabularx}{\textwidth}{@{}Y@{}Y@{}Y@{}}
(d) PU-GAN \cite{li2019pugan} & (e) Dis-PU \cite{li2021dispu} & (f) Ours (discrete)
\end{tabularx}

\par\noindent\rule{1.0\textwidth}{0.4pt}
\vskip 0.1in

\includegraphics[width=\textwidth]{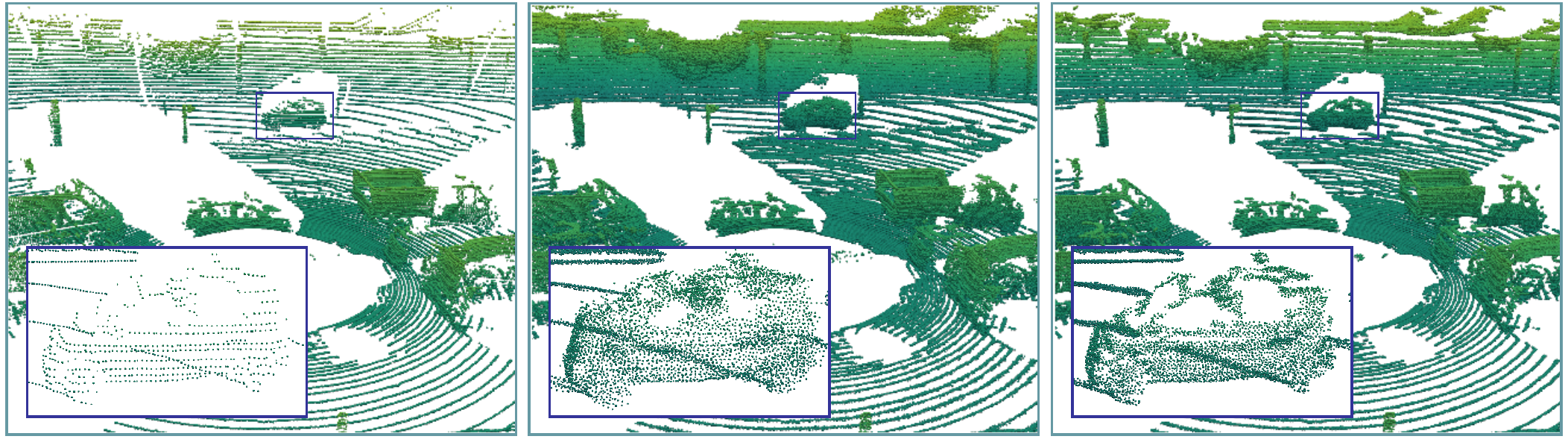}
\begin{tabularx}{\textwidth}{@{}Y@{}Y@{}Y@{}}
(a) Input & (b) PU-GCN \cite{Qian_2021_CVPR} & (c) PUGeo-Net \cite{qian2020pugeo}
\end{tabularx}
\includegraphics[width=\textwidth]{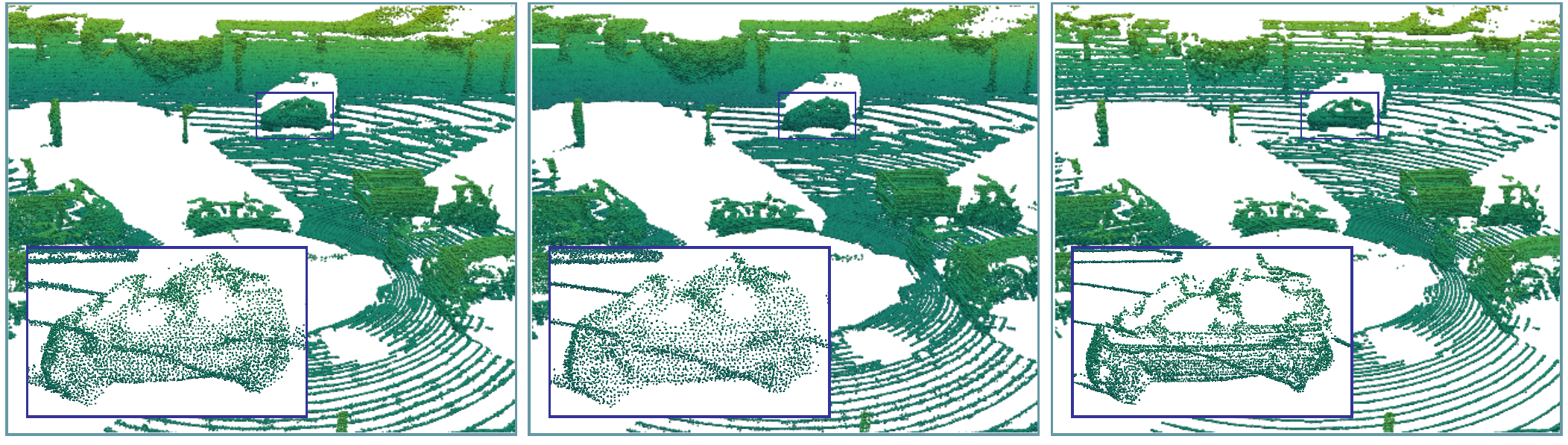}
\begin{tabularx}{\textwidth}{@{}Y@{}Y@{}Y@{}}
(d) PU-GAN \cite{li2019pugan} & (e) Dis-PU \cite{li2021dispu} & (f) Ours (discrete)
\end{tabularx}
\caption{More visual comparisons of upsampling results ($R=4$) on KITTI \cite{Geiger2013IJRR} dataset.}
\label{fig:more-visual-comparison-KITTI}
\end{figure*}

\begin{figure*}[p]
\centering
\includegraphics[width=\linewidth]{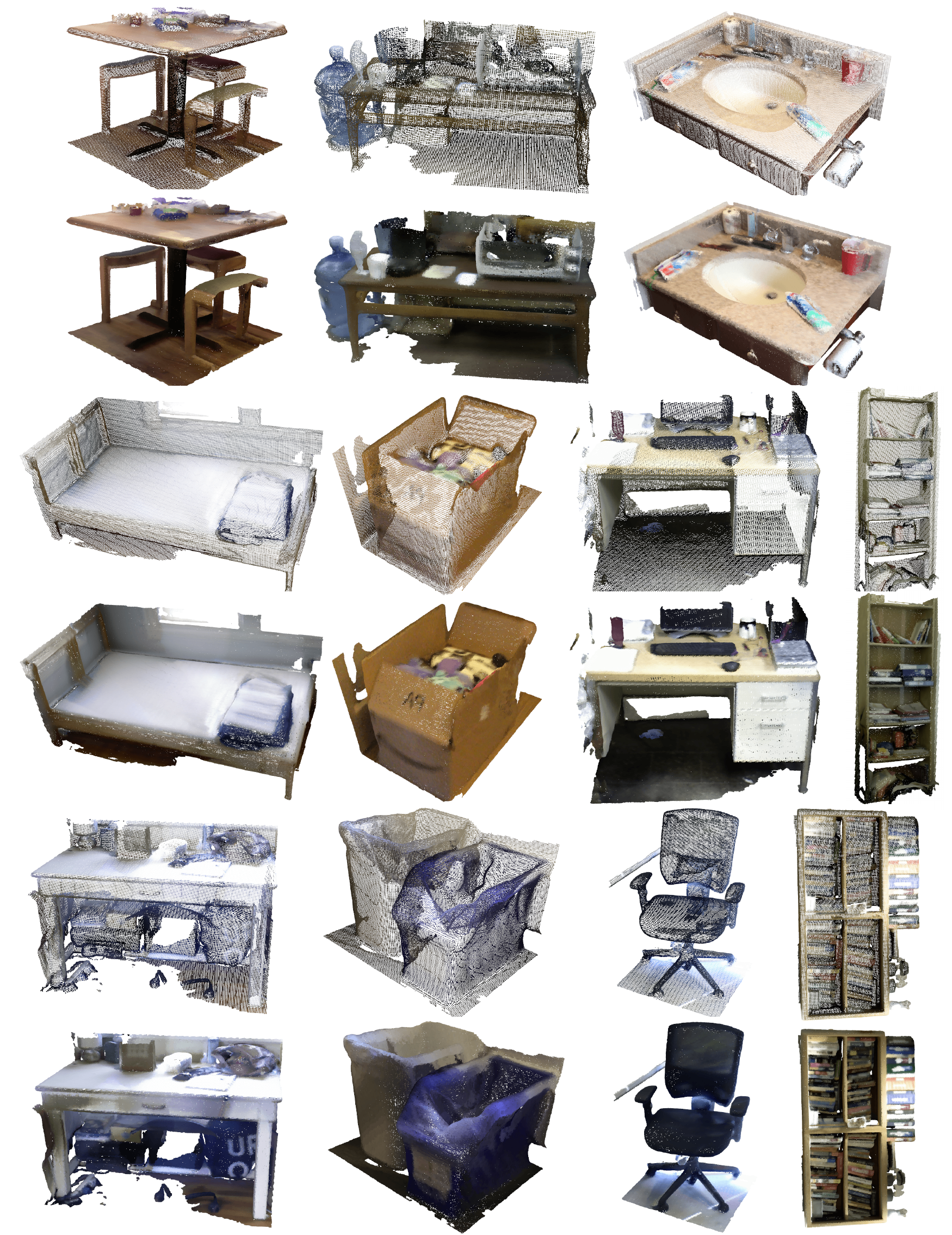}
\caption{
More upsampling results by PU-Flow on different categories (bed, box, desk, bin, chair, shelf).
For better visualization, the color of upsampled point clouds (2nd, 4th, 6th rows) are picked from closest point of input point clouds (1st, 3rd, 5th rows).
}
\label{fig:scanobjectnn-part2-v1}
\end{figure*}



\begin{figure*}[p]
\centering
\ifdefined\GENERALCOMPILE
    \ifdefined\HIGHRESOLUTIONFIGURE
        \includegraphics[width=\linewidth]{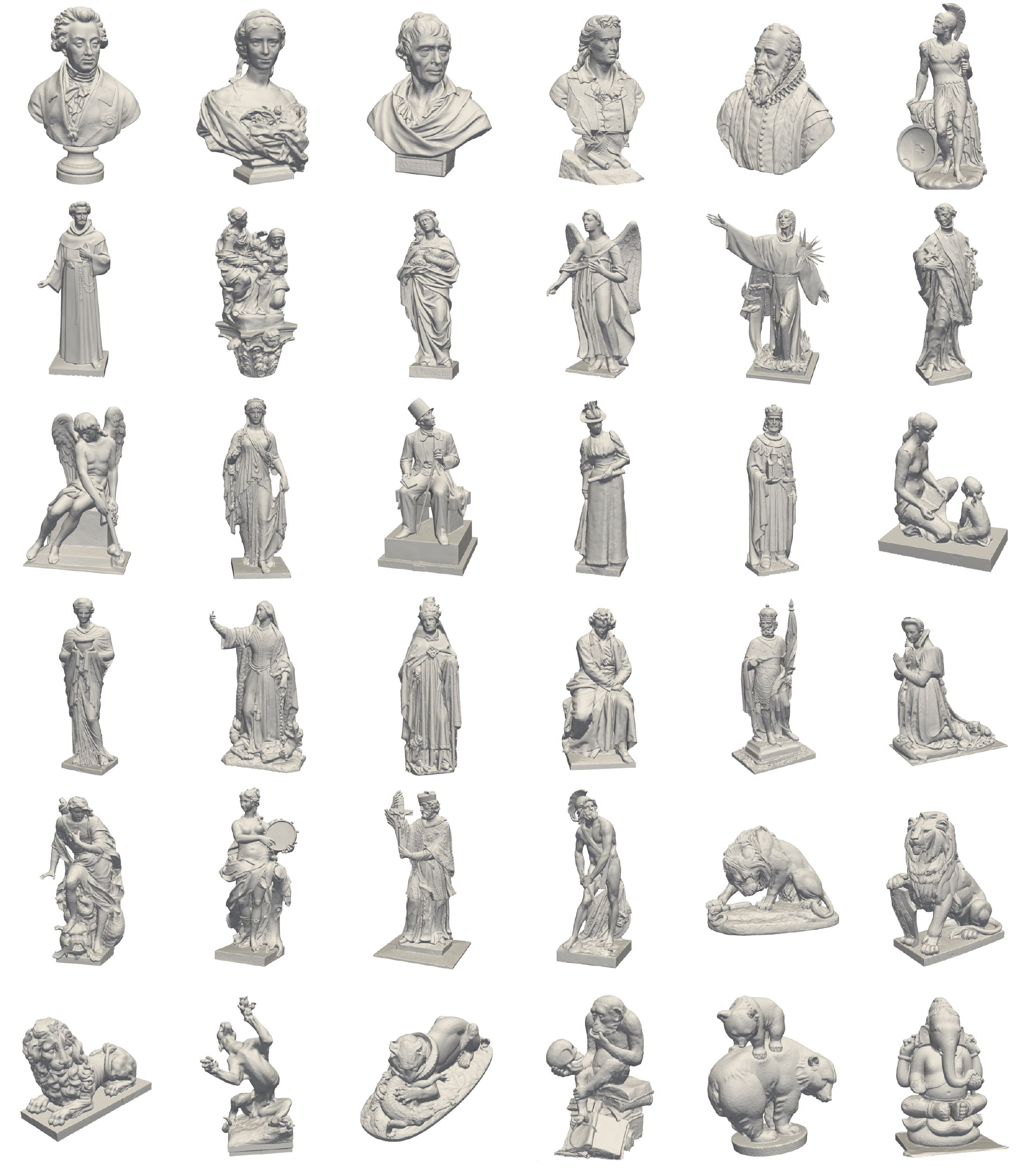}
    \else
        \includegraphics[width=\linewidth]{images/plotting/reduce/dataset-models-reduce.pdf}
    \fi
\fi
\caption{All mesh models in PU36 for evaluation.}
\label{fig:supplement-dataset-models}
\end{figure*}

\fi

\end{document}